\documentclass{article}

\PassOptionsToPackage{numbers, compress}{natbib}

 \usepackage[preprint]{neurips_2020}

\usepackage{microtype}
\usepackage[pdftex]{graphicx}
\usepackage{subfigure}
\usepackage{booktabs} 
\usepackage[utf8]{inputenc} 
\usepackage[T1]{fontenc}    
\usepackage{hyperref}       
\usepackage{url}            
\usepackage{booktabs}       
\usepackage{amsfonts}       
\usepackage{nicefrac}       
\usepackage{microtype}      
\usepackage{algorithm}
\usepackage{algorithmic}
\usepackage{amsmath}
\usepackage{latexsym}
\usepackage{dsfont}
\usepackage{mathrsfs}
\usepackage{amssymb}
\usepackage{amsfonts}
\usepackage{xspace}
\usepackage{color}
\usepackage{amsthm}
\usepackage{balance}
\usepackage{enumitem}
\usepackage{wrapfig}

\newtheorem{theorem}{Theorem}
\newtheorem{lemma}{Lemma}

\newtheorem{definition}{Definition}
\newtheorem{condition}{Condition}

\newcommand{\dA}{\mathds{A}}
\newcommand{\cM}{\mathcal{M}}

\newcommand{\cX}{\mathcal{X}}
\newcommand{\cW}{\mathcal{W}}

\newcommand{\bz}{\boldsymbol{z}}
\newcommand{\bv}{\boldsymbol{v}}
\newcommand{\bc}{\boldsymbol{c}}
\newcommand{\proj}{\mathbf{proj}}

\newcommand{\ours}{\texttt{BMP}}

\newcommand{\bool}{\mathbb{Z}_2}
\newcommand{\real}{\mathbb{R}}
\DeclareMathOperator*{\argmax}{arg\,max}
\DeclareMathOperator*{\argmin}{arg\,min}

\title{Efficient Tensor Decomposition with Boolean Factors}

\author{%
  Sung-En Chang\\
  Northeastern University
  \And
    \hspace{-10mm}
  Xun Zheng\\
  Carnegie Mellon University
  \And 
    \hspace{-10mm}
  Ian E.H. Yen \\
  Carnegie Mellon University
  \AND
  Pradeep Ravikumar \\
   Carnegie Mellon University
      \And 
    Rose Yu \\
     UC San Diego
}

\begin{document}
\maketitle
\begin{abstract}
Tensor decomposition has been extensively used as a tool for  exploratory analysis. Motivated by neuroscience applications, we study tensor decomposition  with Boolean factors. The resulting optimization problem is challenging due to the non-convex objective and the combinatorial constraints. We propose Binary Matching Pursuit (\texttt{BMP}), a novel generalization of the matching pursuit strategy to decompose the tensor efficiently.  \texttt{BMP} iteratively searches for atoms in a greedy fashion. The greedy atom search step is solved efficiently via a MAXCUT-like boolean quadratic program. We prove that \texttt{BMP}  is guaranteed to converge sublinearly to the optimal solution and recover the factors under mild identifiability conditions. Experiments demonstrate the superior performance of our method over baselines  on  synthetic and real  datasets. We also showcase the application of \texttt{BMP}   in quantifying neural interactions underlying high-resolution spatiotemporal  ECoG recordings.


\end{abstract}
\section{Introduction}
\label{sec:intro}
Tensors, as high-order generalizations of matrices, provide concise representation for multi-way data. Tensor decomposition, with direct connections with latent variable modeling \cite{anandkumar2014tensor},  has been a popular tool for exploratory analysis, e.g.  \cite{bahadori2014fast, williams2018unsupervised}. Most tensor decomposition methods assume all the  factors are continuous-valued representing a mixture of all latent components. However, Boolean factors indicating the presence or absence of latent components are  preferred in certain applications such as molecular genetics \cite{rukat2018probabilistic} and clinic medicine \cite{perros2018sustain}. In neuroscience, for instance, given spatiotemporal neural activities,  Boolean factors  can better help us answer the ``when'' and ``where'' questions regarding the underlying brain network  patterns.  This motivates the study of tensor decomposition methods with Boolean factors in this paper. 

The difficulty of  tensor decomposition mainly stems from the fact that  the set of low-rank tensors $\{ \cX| \cX \in \real^{d_1\times d_2\times d_3}, \text{rank}(\cX) \leq R \}$ is non-convex and in general not closed. As such, the Maximum Likelihood Estimator (MLE) objective is non-convex and the best rank-R approximation of a tensor may not exist \cite{de2008tensor}. This difficulty  is magnified by the combinatorial constraint of Boolean factors. To bypass the MLE objective, \cite{tung2014spectral, anandkumar2014tensor, jaffe2018learning,ge2017optimization} proposed a method of moments estimator, which achieves global guarantees in the average case. However, they rely on the strong distributional assumptions on the latent factors and can be unstable in  model mis-specification cases.   \cite{tomioka2011statistical, yuan2016tensor,  wimalawarne2018efficient} propose to use  nuclear norm as a convex surrogate for the rank constraints, but can be computational challenging for large-scale problems.

To tackle the Boolean constraint, \cite{slawski2013matrix} considered the noiseless case and proposed a geometric algorithm, but their method has exponential complexity in the rank of the decomposition. \cite{yen2017latent} improved the solver using  convex relaxation that achieves  linear sample complexity w.r.t the rank and the dimensions. Our setting is a special case of Boolean tensor decomposition \cite{metzler2015clustering, perros2018sustain} where the input tensor is also Boolean. \cite{metzler2015clustering, perros2018sustain} proposed algorithms based on alternating least square (ALS)  followed by rounding heuristics.  However, ALS does not perform well in the presence of highly noisy measurements.  \cite{rukat2018probabilistic} studied a Bayesian version of the problem and apply Monte Carlo sampling. The theoretical behavior of these methods are not well understood. 

We provide an efficient solution for learning tensor decomposition with Boolean factors. Using atomic norm, we cast the non-convex tensor decomposition problem as a convex program with sparsity constraints, which enjoys tractable relaxations. 
We propose a novel algorithm, Boolean Matching Pursuit (\ours{}), to search for  atoms of the steepest descent  iteratively. Our algorithm enjoys strong theoretical guarantees. It can recover the parameters exactly under identifiability conditions with sublinear convergence.   The sample complexity scales only linearly with the rank of the tensor.  We validated the superior performance of \ours{} on synthetic and neural recordings.  In summary, our contributions include:
\begin{itemize}[noitemsep]
    \item We study a novel tensor decomposition model with Boolean factors, which is particularly suitable for exploratory analysis of discretized spatiotemporal data.
    \item We formulate the non-convex problem as an atomic-norm regularized convex program and  propose a fast  algorithm Binary Matching Pursuit (\ours{}) to solve the problem efficiently.
    \item Our algorithm is guaranteed to converge sub-linearly to the optimal solution, with run-time and sample complexity only linear in the number of atoms.
    \item We experiment extensively on synthetic and real-world ECoG datasets and observe superior performance for denoising and completion tasks. Our algorithm also uncovers the interesting neural mechanism underlying consciousness in brain computer interface (BCI).
\end{itemize}

\newcommand{\T}[1]{\mathcal{#1}}
\newcommand{\M}[1]{\mathbf{#1}}
\newcommand{\V}[1]{\boldsymbol{#1}}

\section{Related Work}
\label{sec:relate}
\paragraph{Tensor decomposition.}
Tensor decomposition has been the subject of extensive study; please see the review paper by \cite{kolda2009tensor} and references therein. Most tensor factorization work focuses on extracting high-order structure with continues-value factors. For instance, \cite{jain2014provable,sharan2017orthogonalized} proposed orthogonalized ALS to decompose a tensor alternatively but are only limited to orthogonal factors.  \cite{gandy2011tensor, tomioka2013convex, williams2018unsupervised} developed nuclear norm regularization as a convex surrogate and solve the problem using  alternating direction method of multipliers (ADMM). But  can suffer from high computational costs. \cite{bahadori2014fast} designed a non-convex solver based on a greedy algorithm  and demonstrate significant speedup. There has also been work on Boolean tensor decomposition where the input tensor has Boolean values \cite{miettinen2011boolean, rai2015scalable}, which is different from our problem where the learned factors are Boolean. \cite{metzler2015clustering, perros2018sustain} proposed algorithms based on alternating least square (ALS) with rounding heuristics.  To the best of our knowledge, our work is the first algorithm for tensor decomposition with Boolean factors with theoretical guarantees.

\paragraph{Boolean constrained latent variable model.}
Latent variable model with Boolean constraints is also known as latent feature model (LFM) \cite{ghahramani2006infinite} in statistical learning, where each observation is associated with a set of real-valued latent features and Boolean vector indicating the presence/absence of the features.  For the parametric version of the model,  \cite{tung2014spectral, anandkumar2014tensor, jaffe2018learning} propose to use spectral methods to estimate the moments of the distribution at different orders, but can suffer from high sample complexity. Under certain identifiability condition, \cite{slawski2013matrix} proposed a convex optimization algorithm  by selecting a maximal affine independent subset. However, the selection process in their algorithm has an exponential computational complexity.  Perhaps the work that is most related to ours is \cite{yen2017latent} in which a convex estimator for matrix latent feature models which under certain identifiability conditions achieves a linear sample complexity. LFMs also bear affinity to sparse dictionary learning \cite{kreutz2003dictionary} whereas the representations are real-valued instead of Boolean.


\vspace{-3mm}

\paragraph{Preliminary}
Across the paper, we use calligraphy font for tensors, such as $\T{X},\T{Y}$, bold uppercase  for matrices, such as $\M{X},\M{Y}$, and bold lowercase  for vectors, such as $\V{x},\V{y}$.
For easy of illustration, we use order-$3$ tensor throughout the paper. Our results directly generalize to high-order cases.

\textit{Mode-$n$ Unfolding}:
For an order-$3$ tensor $ \mathcal{X} \in \mathbb{R}^{d_1 \times d_2 \times d_3} $, a  mode-$n$ unfolding is to matricize a tensor along a particular mode $ \M{X} = \mathrm{unfold}_{n}(\mathcal{X}) $ with $d_n$ rows and $\prod_{i}d_i/d_n$ columns. The mode-$n$ refold  $ \mathcal{X} = \mathrm{refold}_n(\M{X}) $ is the reverse operation. 
The indexing follows the convention in \cite{kolda2009tensor}.  

\textit{Tensor Rank}:
The rank of a tensor is the minimum number of rank-1 components it contains: \\
$\text{rank}_{\otimes}(\T{X})=:\{\min R| \T{X})=\sum_{r=1}^R \V{u}_r \otimes \V{v}_r\otimes \V{w}_r \}$.
Multilinear rank  is a tuple $(R_1, R_2, R_3)$ such that $R_n = \text{rank}(\text{unfold}_n(\T{X}) )$. We have $R_n \leq \min(\text{rank}_{\otimes}(\T{X}), d_n)$.




\section{Tensor Decomposition with Boolean Factors}
\label{sec:method}
\subsection{Boolean Canonical Polyadic Decomposition}
Consider the following tensor decomposition model for an order-$N (N=3)$  tensor $\T{X} \in \real^{d_1\times d_2 \times d_3}$:
\begin{equation}
    \T{X} = \sum_{n=1}^{N} \sum_{r=1}^R  \V{z}^r_n \otimes \V{u}^r_n  \otimes \V{w}^r_n  + \T{E}
    \label{eqn:model}
\end{equation}
%
where one of the factors is Boolean and the rest are continuous-valued.  The noise tensor $\T{E}$ has the same size as $\T{X}$ with i.i.d Gaussian entries of zero mean and variance $\sigma^2$. The subscript $n$ indicates the mode with Boolean factors. The dimensions of the factors change accordingly. For instance,  if the first mode factors are Boolean, $\V{z}^r_1 \in \bool^{d_1}$, $\V{u}^r_1 \in \real^{d_2}$ and $\V{w}^r_1 \in \real^{d_3}$. 

%
The tensor decomposition model  in \eqref{eqn:model}  generalizes the latent feature model \cite{ghahramani2006infinite} to high-order tensors. Continues-valued factors represent  latent features at every mode and Boolean factors indicate the presence/absence of these features. It resembles latent  mixture model  \cite{tomioka2013convex} which models a tensor as  a mixture of latent  tensors across modes.  Latent mixture model assumes all terms to be  continuous-valued, whereas our model contains Boolean factors. We name the model \eqref{eqn:model} Boolean Canonical Polyadic Decomposition.

\subsection{Atomic Norm Regularized Convex Program}
The estimation problem of the model in \eqref{eqn:model} is generally intractable due to non-convex  optimization objective and   combinatorial Boolean constraints.  To make the learning tractable, we note that a tensor can be expressed as  a linear combination of rank-1 tensors, or atoms.   Define  a mode-$n$  atomic set as  $\mathds{A}_n :=\{\T{M}| \T{M} = \V{z}^r_n\otimes\V{u}^r_n\otimes \V{w}^r_n \}$. The union set $\mathds{A}=\bigcup_n \mathds{A}_n $ contains all the rank-1 tensor with arbitrary Boolean factors, whose size is $\bar{K} \leq 3R$. Then the tensor decomposition problem in  ~\eqref{eqn:model} can be  reformulated as   a sparsity constrained convex program, which enjoys tractable relaxations.   In particular, given an observation $ \mathcal{X} \in \real^{d_1 \times d_2 \times d_3}$, we would like to find a sparse representation of $ \mathcal{X} $ in the subspace of atoms $\mathds{A} $. We can write down the  tensor decomposition problem over the atom set: 
\begin{align}
\min_{ \boldsymbol{c}, \mathcal{W}  }  & \quad F(\cW):= \frac{1}{2} \| \mathcal{X} -  \T{W} \|_F^2  \quad
\quad  \mathrm{s.t.} \quad   \T{W}=\sum_{\mathcal{M}_k \in \mathds{A}} \V{c}_k \mathcal{M}_k, \ \| \boldsymbol{c} \|_0 \leq \bar{K}
 \label{eqn:problem}
\end{align}
where $ \|\mathcal{W}\|_F = \sqrt{ \sum_{i_1, \dotsc, i_L } \mathcal{W}_{i_1, \dotsc, i_L}^2  } $  is the tensor Frobenius norm, and $\V{c}$ is a vector of coefficients.

The non-convexity of $ \ell_0 $ norm as well as the large  number of atoms in $ \mathds{A} $ make the problem in ~\eqref{eqn:problem}  difficult.  We can use the  $ \ell_1 $ relaxation as a convex surrogate to $ \ell_0 $ norm, we can then utilize the notion of an \textit{atomic norm}~\cite{chandrasekaran2012convex} as a key technical device to remedy this problem, which leads to the following  tractable  formulation given the  atomic norm of  a tensor $\T{W}$:
\begin{align}
\min_{ \mathcal{W}  }  \ F(\mathcal{W}) + \| \mathcal{W} \|_{\mathds{A}}, \quad 
\| \mathcal{W} \|_{\mathds{A}}  = \big\{ \inf \ \| \boldsymbol{c} \|_1 : 
\mathcal{W} = \sum_{\mathcal{M}_k \in \mathds{A}} \V{c}_k \mathcal{M}_k,  \T{M} = \V{z}^r_n\otimes\V{u}^r_n\otimes \V{w}^r_n  \big\}
\label{eqn:compact_problem}
\end{align}
which is a convex program with an atomic norm regularizer defined over the subspace of rank-1 tensors. The atomic norm regularized problem \eqref{eqn:compact_problem}  is  still  difficult as the size of the atomic set grows exponentially with the order of the tensor.  To avoid exhaustive search in the atomic set, we develop a fast and easy to implement optimization algorithm based on matching pursuit.

\subsection{Background in Matching Pursuit}
Matching pursuit (MP) ~\cite{mallat1993matching,tropp2006algorithms} is a sparse approximation procedure that aims to find the ``best match'' of the data onto a set of atoms.  Matching pursuit was initially developed for the continuous-valued  vector inputs. Specifically, given an input vector $\V{x} \in  \real^d$, MP approximates $\V{x}$ using a linear combination of atoms $\V{x} \approx \sum_k \V{c}_k \V{m}_k, \V{m}_k \in \real^d$. At each iteration, MP greedily searches for the atom  $\V{m}_k$ that maximizes the inner product with  the residual and update the weights $\V{c}$ either incrementally (greedy MP) or fully (orthogonal MP). In this paper, we generalize matching pursuit to solve tensor decomposition problems with Boolean constraints.  The main difficulty of the generalization is the greedy atom search in the set of atoms, as each atom in our setting is a rank-1 tensor with a mixture of Boolean and continuous-valued components. 

\begin{minipage}[t]{0.48\textwidth}
\begin{algorithm}[H]
\caption{\texttt{BooleanMatchingPursuit}}\label{al1}
\begin{algorithmic}[1]
\STATE \textbf{Input:} Tensor $ \mathcal{X} $ \\
\STATE \textbf{Output:} Tensor $ \mathcal{W} $
\STATE Initialize $ \mathcal{W}^{0} \gets 0$ , an active set $\mathds{A} \gets \emptyset$ 
\FOR{$ k= 0, \dotsc, K-1 $}
\STATE  $ \T{M}_k \gets \mathrm{\texttt{GreedyAtomSearch}}(\mathcal{X}, \mathcal{W}^{k}) $
\STATE Add to the active set $ \mathds{A} \gets \mathds{A} \cup \{\T{M}_k\} $
\STATE Adjust coefficients $ \boldsymbol{c}$ by solving \eqref{adjustment}
\STATE Reconstruct $ \mathcal{W}^{k+1} \gets \sum_{\T{M}_k\in \mathds{A}} \V{c}_k \T{M}_k $
\ENDFOR
\end{algorithmic}
\end{algorithm}
\end{minipage}
\hfill
\begin{minipage}[t]{0.48\textwidth}
\begin{algorithm}[H]
\caption{\texttt{GreedyAtomSearch}}\label{al2}
\begin{algorithmic}[1]
\STATE \textbf{Input:} Tensor $ \mathcal{X}, \mathcal{W} $
\STATE \textbf{Output:} Rank-1 tensor $ \T{M} $
\FOR{each  mode $n=1\,\dots, N$} 
\STATE Gradient $ G \gets \mathrm{unfold}_n (\nabla F(\mathcal{W} )) $
\STATE Quadratic term $ C \gets - G  G^T $
\STATE Solve MAXCUT $ \bz \gets \mathrm{\texttt{MaxCut}} (C) $
\STATE Unit vector $ \bv \propto  G^T \bz $
\STATE Candidate atom $ \T{M}_n \gets \text{refold}(\V{z}_n\otimes \V{v})  $
\ENDFOR 
\STATE 
$
\T{M} \gets \argmin_{ \T{M}_n }  \ \big\langle \nabla F(\mathcal{W}), \mathcal{M}_n \big\rangle \nonumber
$.
\end{algorithmic}
\end{algorithm}
\end{minipage}

\section{Boolean Matching Pursuit (BMP)}
\label{sec:alg}
 We introduce Boolean Matching Pursuit (\texttt{BMP}), a generalization of orthogonal matching pursuit algorithm to Boolean tensors. We exploit the structure of the Boolean constraint and propose an efficient MAXCUT-like Boolean quadratic solver to greedily search for the atoms.  

The high level mechanism of our proposed \texttt{BMP} algorithm is the same as MP. As detailed in  Algorithm~\ref{al1},  we maintain an active set of atoms $\mathds{A}$ for the selected atoms. The solution is approximated by a weighted combination of atoms $\T{W} = \sum_{\T{M}_k\in \mathds{A}} \V{c}_k \T{M}_k $ iteratively. Every time, the subroutine greedily selects an atom, add this atom to the active set, followed by a full adjustment of weights to minimize the approximation error. 

\subsection{Greedy Atom Search}
We propose an efficient search procedure for atoms with Boolean factors.  At iteration $ k $, we  greedily search for a rank-1 tensor $ \mathcal{M}=\V{z}_n\otimes \V{u}_n\otimes \V{w}_n$ that corresponds  to the steepest descent (also maximum inner product) direction in the atom set across all modes. 
\begin{align}\label{greedy_step}
 \min_n\min_{\mathcal{M} \in \mathds{A}_n } \  \big\langle \nabla F(\mathcal{W}^{k}), \mathcal{M} \big\rangle.
\end{align}
Our algorithm is ``greedy''  both across the  modes and within each model of the unfolded tensor.  A key algorithmic contribution of this work is a novel solution based on a MAXCUT like Boolean quadratic program to optimize the inner minimization problem efficiently.

Without loss of generality, assume the inner minimization  solves for mode-1,  we can unfold $ \nabla F(\mathcal{W}) $  into a $ d_1  \times d_2d_3 $ matrix. Then the  minimization problem in ~\eqref{greedy_step} can be written as 
\begin{align}
\min_{
\bz\in\bool^{d_1},\bv\in \real^{d_2d_3}
} 
\ \big\langle \text{unfold}_1 (\nabla F(\mathcal{W}^{k})), \bz \bv^T \big\rangle.
\end{align}

Where the vector $\bv = \text{vec}(\V{u}\otimes \V{w})$ is normalized and  lies in the space of continuous-valued unit vectors, when fixing $\bz$ and minimizing w.r.t. $\bv$,  we have 
$$
\bv^*(\bz)= \frac{\text{unfold}_1 (\nabla F(\mathcal{W}^{k}))^T \bz}{\| \text{unfold}_1(\nabla F(\mathcal{W}^{k}))^T \bz\|}
$$
Therefore, the joint minimization problem w.r.t. $(\bz,\bv)$ is equivalent to finding $\V{z}^*$ such that 
\begin{align}\label{maxcut}
\bz^* = \argmin_{\bz \in \bool^{d_1} } \  \bz^T \underbrace{ 
\nabla F(\mathcal{W}^{k}) \nabla F(\mathcal{W}^{k})^T 
}_{-C} \bz 
& = \argmax_{\bz\in \bool^{d_1} }\   \bz^T C\bz.
\end{align}
which can be solved  using a Boolean quadratic solver efficiently.

\subsection{Boolean Quadratic Program}
\newcommand{\ones}{\boldsymbol{1}}
With change of variables,  the  problem in Eqn \eqref{maxcut} is a MAXCUT-like problem, which enjoys constant approximation guarantees  \cite{yen2017latent}. Specifically, define a vector $ \V{y} = 2\V{z} - 1 $, then $ \boldsymbol{y} \in \{-1,1\}^{d_1} $. 
Augmented with a dummy variable $ y_0 \in \{-1,1\} $, the problem can be rewritten as 
\begin{align}
\max_{[y_0; \boldsymbol{y}] \in \{-1,1\}^{d_1+1}  }  \ \frac{1}{4} 
\begin{bmatrix}
y_0 \\
\boldsymbol{y}
\end{bmatrix}^T
\underbrace{
\begin{bmatrix}
\ones^T C \ones & \ones^T C \\
C \ones & C 
\end{bmatrix}
}_{\widetilde{C}}
\begin{bmatrix}
y_0 \\
\boldsymbol{y}
\end{bmatrix},
\end{align}
which is now in a MAXCUT-like formulation. 
In general, even if the quadratic factor is positive definite, i.e., $ \widetilde{C} \succ 0 $, the decision version of the problem is still NP-complete~\cite{garey1979computers}. 
However, there exist semidefinite programming (SDP) relaxations that has constant factor approximation guarantees for the following problem:
\begin{align}\label{sdp}
\max_{ Y \succeq 0 }  & \quad \big\langle \widetilde{C}, Y \big \rangle \quad
\mathrm{s.t.}  \quad \mathrm{diag} (Y) = \ones,
\end{align}
Rounding of the solution to the above SDP problem guarantees a $ 3/5 $-approximation~\cite{nesterov1997quality}. 

Indeed, although the polynomial time complexity of SDPs is already a big saving compared to NP-completeness, it is still impractical to employ a general SDP solver as the subroutine. Fortunately, unlike general SDPs, the SDP of the form \eqref{sdp} has specialized solver ~\cite{wang2017mixing}, whose time complexity only depends linearly on the number of non-zeros in $\widetilde{C} $.   This subroutine is described in  Algorithm~\ref{al2}.


With an updated set of atoms $\mathds{A}$, we can  adjust the atom weights to reflect changes in the atom set. Fixing the active set, the weight adjustment is a simple least-square problem in terms of $ \boldsymbol{c} $ as:
\begin{align}\label{adjustment}
\min_{\boldsymbol{c} \in \mathbb{R}^{|\mathds{A}|}  } & \quad \frac{1}{2} \| \mathcal{X} - \sum_{\mathcal{M}_k \in \mathds{A}} \V{c}_{k} \mathcal{M}_k \|_F^2 ,
\end{align}
with a closed-form solution:
$\bc = (\M{M}^top \M{M})^{-1} \M{M}^\top \mathrm{vec}(\mathcal{X}) $
where $\M{M}$ is a matrix of size $d_1d_2d_3\times \bar{K}$, each column of which  is a flattened atom  $ \mathrm{vec} (\T{M}) $ in the active set. 

\section{Theoretical Analysis}
\label{sec:theory}
\subsection{Convergence Analysis}
An important property of the Algorithm \ref{al1} is that the number of atoms comprising the output tensor $\T{W}^{K}$ is equal to the number of iterations $K$, which leads to a trade-off between optimization error and computation. Assume the optimal solution $\T{W}^* \in \mathds{A}$ has   $\bar{K}$ number of atoms. By running \texttt{BMP} for $\Omega(\bar{K}/\epsilon)$ iterations, one can achieve a sublinear convergence to the  optimal solution.

\begin{theorem}\label{thm:convergence1}
Let $\cW^*$ be the optimal solution of the problem \eqref{eqn:compact_problem}, and $ \{\cW^{k}\}_{k=1}^K$ be the sequence of iterates produced by the BMP Algorithm \ref{al1}. Assume the loss function $F$ is $\beta$-smooth, then
\begin{equation}\label{convergence1}
F(\cW^{k})-F(\cW^*)\leq \frac{2\beta\|\cW^*\|_{\dA}^2}{\mu^2}\left(\frac{1}{k}\right).
\end{equation}
for any $k\in[K]$ and $\mu$ is the approximation ratio for the SDP solver.
\end{theorem}

\textbf{Remark}: A key ingredient of our analysis is the $\mu=3/5$ constant-approximation guarantee in the greedy step given by the SDP-based MAXCUT solver, where we utilize the following guarantee for the atom $\T{M}^*$ picked by Algorithm \ref{al2}:
\begin{equation}\label{greedy_guarantee}
\langle \nabla F(\cW), \cM^* \rangle \leq \mu \left(\min_{\cM\in\dA} \; \langle \nabla F(\cW), \cM \rangle \right).
\end{equation}


Theorem \ref{thm:convergence1} only establishes an error bound relative to the atomic norm of the optimal solution $\|\cW^*\|_{\dA}$. If the loss function $F$ is $\gamma$-strongly convex w.r.t the support set,  we have the following additional result that bounds the optimization error directly in the ratio of number of atoms $\bar{K}/K$:
\begin{theorem}  
Assume the loss function $F$ is  $\gamma$-strongly convex w.r.t. the support set. After running $K$ iterations of the BMP Algorithm \ref{al1}, the solution $\T{W}^K$  satisfies
\begin{equation}\label{convergence2}
F(\cW^{K})-F(\cW^*) \leq \frac{2\beta\|\cX\|_F^2}{\gamma\mu^2}\left(\frac{\bar{K}}{K}\right)
\end{equation}
\end{theorem}
which shows linear scaling in terms of the ratio of number of atoms $O(\frac{\bar{K}}{K})$, demonstrating the trade-off between the sparsity of the problem and the computation complexity.
\subsection{Identifiability and Parameter Recovery}
\vspace{-1mm}
Identifiability is of great importance to applications, where one might be interested in interpreting the latent factors. We provide the conditions for single latent tensor where the results directly apply to the mixture model in \eqref{eqn:model}. A tensor decomposition $\T{W}= \sum_{r=1}^R  \V{z}_r \otimes \V{u}_r \otimes \V{w}_r$ is unique if for any other decomposition, $\T{W}= \sum_{r=1}^R  \V{z}_r' \otimes \V{u}_r' \otimes \V{w}_r'$, there exists a permutation $\sigma$ such that 
\[ \V{z}_r \otimes \V{u}_r \otimes \V{w}_r = \V{z}_{\sigma(r)}' \otimes \V{u}_{\sigma(r)}' \otimes \V{w}_{\sigma(r)}'\]
When tensor $\T{W}$ has a unique decomposition, the set of factors $\V{z}$ , $\V{u}$ and $\V{w}$ are identifiable, up to scalars.  Let $\M{Z}$ be the matrix containing $\V{z}_r$ as its columns. Similarly for matrices $\M{U}$ and $\M{W}$.
The following Kruskal's condition \cite{kruskal1977three} guarantees the uniqueness of generic tensor decomposition:
\begin{condition}[Kruskal]\label{thm:kruskal}
An order-3 rank-R
tensor of dimension $d_1\times d_2\times d_3$ is unique if
\[R \leq \frac{1}{2}\left(\text{krank}(\M{Z}) + \text{krank}(\M{U}) + \text{krank}(\M{W}) -2 \right)\]
\end{condition}
\vspace{-2mm}
where $\text{krank}(\cdot)$ is the largest value of a matrix  such that every subset of columns of the matrix is linearly independent. It is also easy to see that $\text{krank}(\M{Z}) \leq \text{rank}(\M{Z})$ for any $\M{Z}$.
%
%
To guarantee uniqueness of the Boolean factors $\M{Z}$, we need  additional conditions on its column vectors. 
\begin{condition}[Rigidity] 
 \label{thm:rigidity}
The tensor decomposition problem has unique Boolean factors  if any  non-trivial combinations of the column vectors $\{\V{z}_r\}$ would lead to a non-Boolean vector:
 \[ \forall \V{c}\neq \V{0}, \quad  \V{c}^\top \V{Z} \in \bool \iff \V{c} \in \{ \V{e}_i\}\]
\end{condition}
which means the linear subspace of $\V{Z}$ does not contain any other Boolean vectors that are not already in $\M{Z}^n$. This identifiability condition is in nature  similar to \cite{slawski2013matrix}. The following theorem guarantees the exact recovery of the latent factors from tensor decomposition. 
\begin{theorem}[Exact Recovery]
Let $\T{X}\in \real^{d_1\times d_2\times d_3}$ be a tensor $\T{X}=\sum_{r=1}^R  \V{z}^r\otimes \V{u}^r \otimes \V{w}^r$. Under the identifiability conditions, the algorithm can recover the atom parameters   $\{\V{z}^r, \V{u}^r, \V{w}^r\}$ exactly. 
\end{theorem}

Note that the above results are stronger than matrix case as matrix factors are not uniquely defined due to invariance under rotations. The guarantee is  deterministic for the noiseless setting. 

We also analyze the statistical performance of the estimator for Problem \eqref{eqn:compact_problem}. Using the notion of restricted strongly convexity \cite{negahban2012unified}, the following theorem guarantees the sample complexity under the Gaussian noise, in which we leverage the properties of the atomic norm.
\begin{theorem}[Sample Complexity]
Assume the true model $\T{W}^* \in \mathds{A}$, and the loss function satisfies the restricted strongly convex condition. There exists a universal constant $c_1$ such that if we choose the regularization parameter $\lambda_S = \sigma \sqrt{\frac{d_1+d_2+d_3}{S}}$, the statistical estimator $\hat{\T{W}}$ satisfies:
\[   \|\hat{\T{W} }  - \T{W}^* \|_F^2   \leq \big(  O_p \big(c_1 \frac{\sigma^2 R(d_1 + d_2+ d_3)}{S} \big) \]
where $S$ is the number of samples and $\sigma^2$ is the noise variance. 
\end{theorem}

\textbf{Computational Complexity \quad} Our algorithm \texttt{BMP} is efficient and easy to implement. Assuming $d_1 \geq d_2 \geq d_3$, the  greedy atom search step involves solving continuous-valued factors in $\mathcal{O} (d_2d_3)$ and an efficient Boolean quadratic program whose complexity is  $\mathcal{O} (d_1^2)$ at each mode. At $k$-th iteration, the least square step can be solved in    $\mathcal{O} (kd_1d_2d_3)$ if we maintain a QR decomposition of $\M{M}$, though faster solution is possible if $\M{M}$ is highly structured \cite{tropp2007signal}.
\vspace{-3mm}

\begin{figure*}[t]
\centering
\subfigure{ \includegraphics[width=0.30\textwidth]{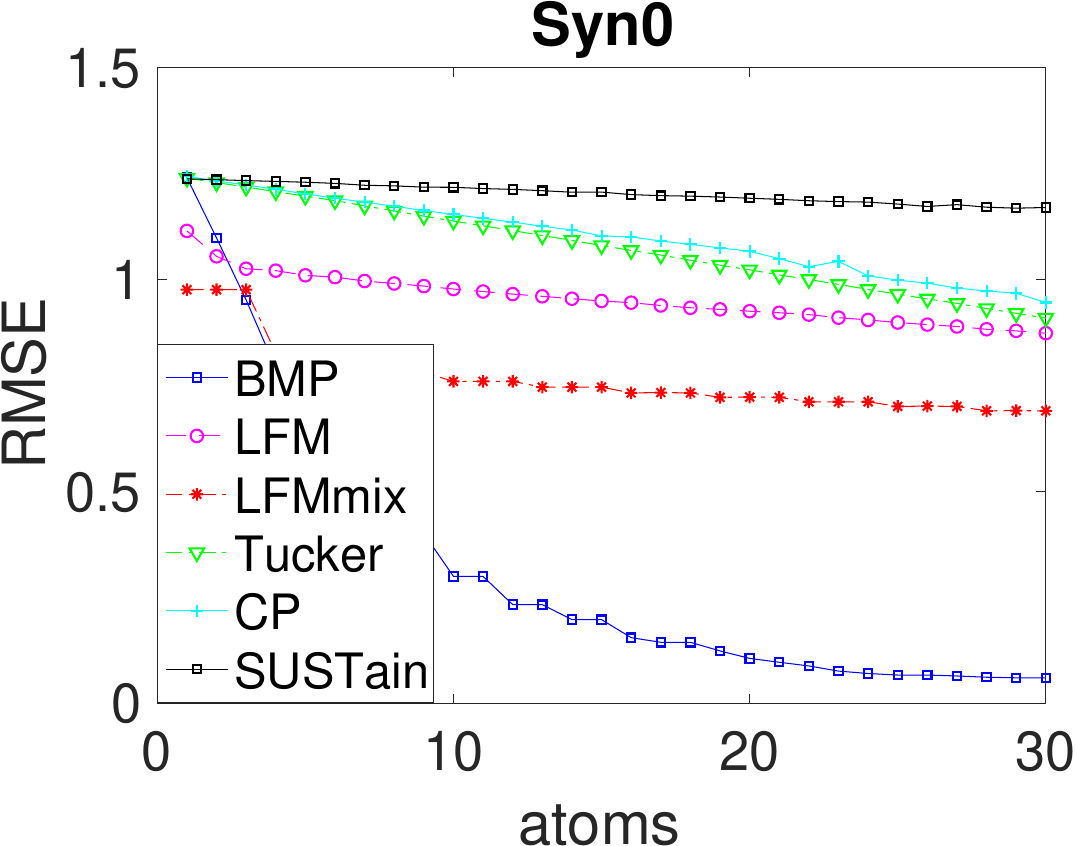}  }
\subfigure{ \includegraphics[width=0.30\textwidth]{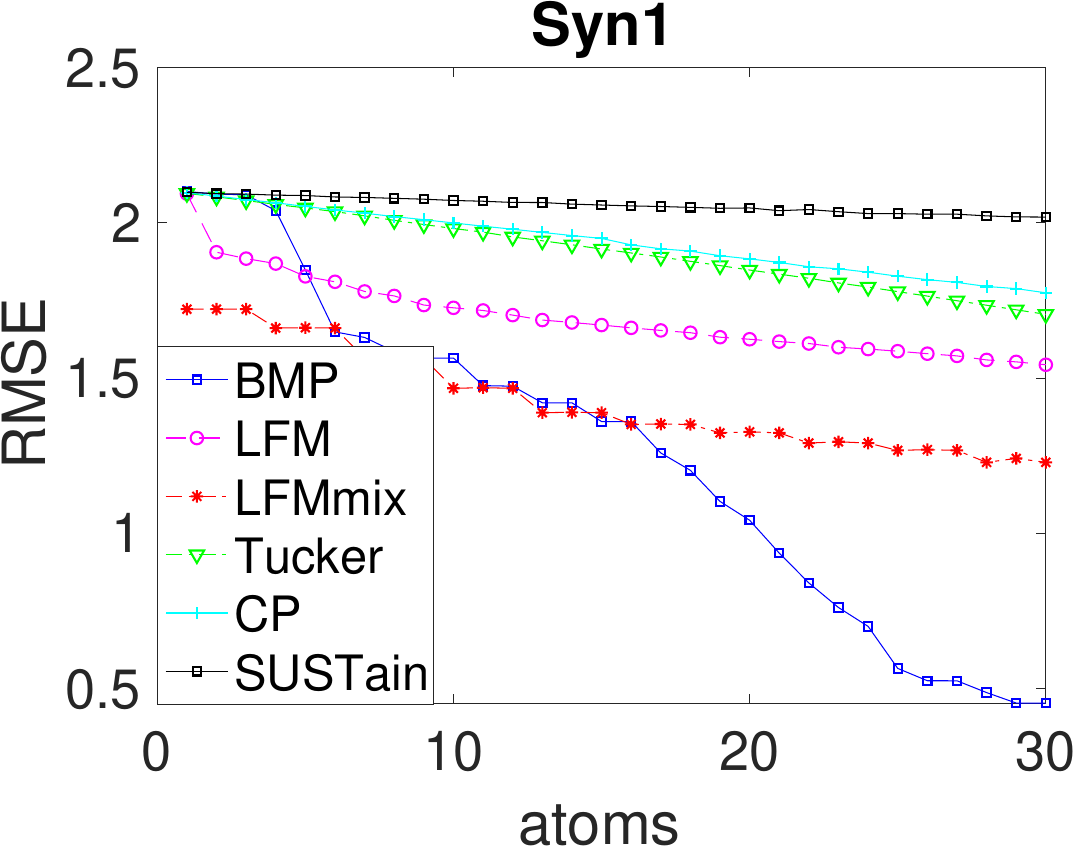}  }
\subfigure{ \includegraphics[width=0.30\textwidth]{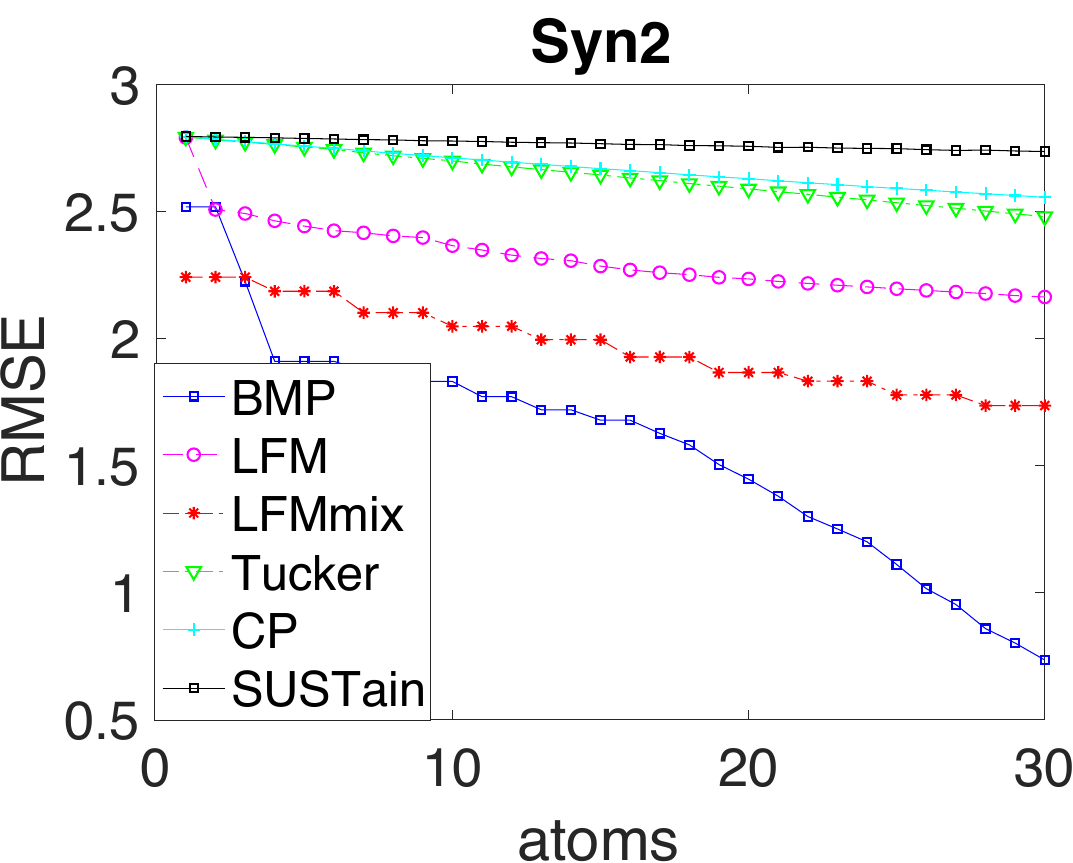}  }\\
\vspace{-4mm}
\subfigure{ \includegraphics[width=0.30\textwidth]{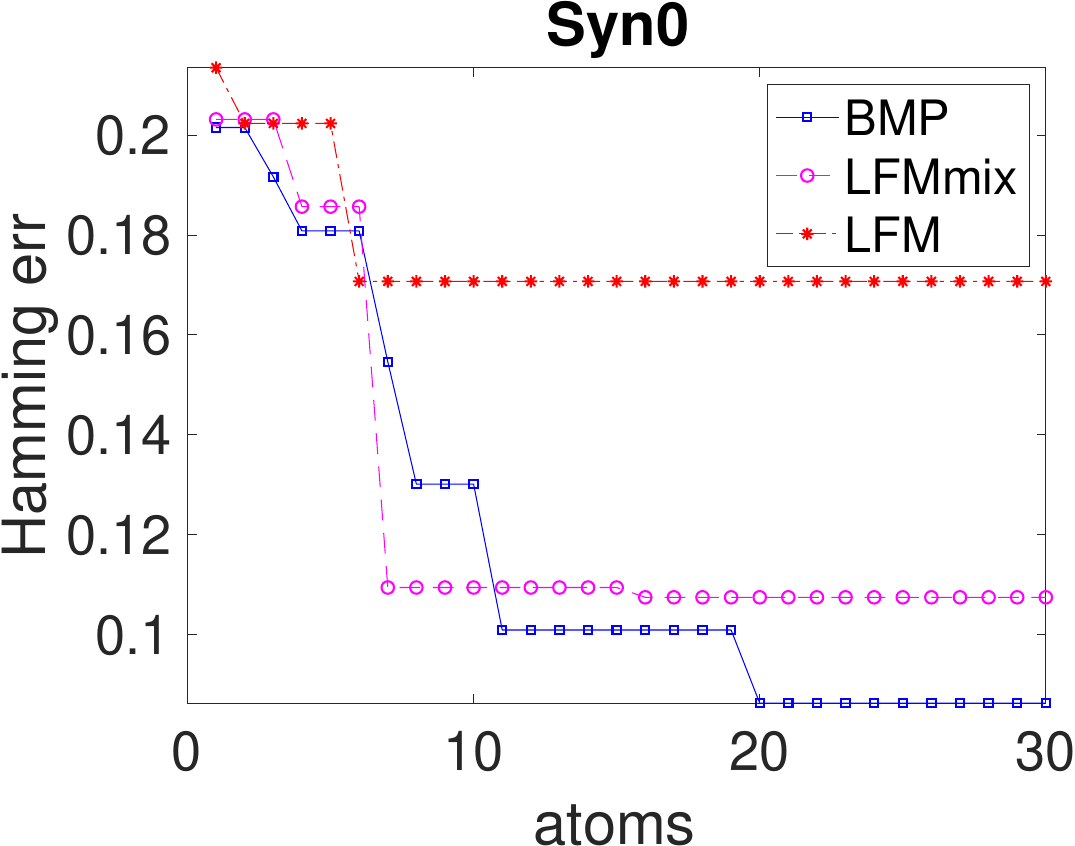}  }
\subfigure{ \includegraphics[width=0.30\textwidth]{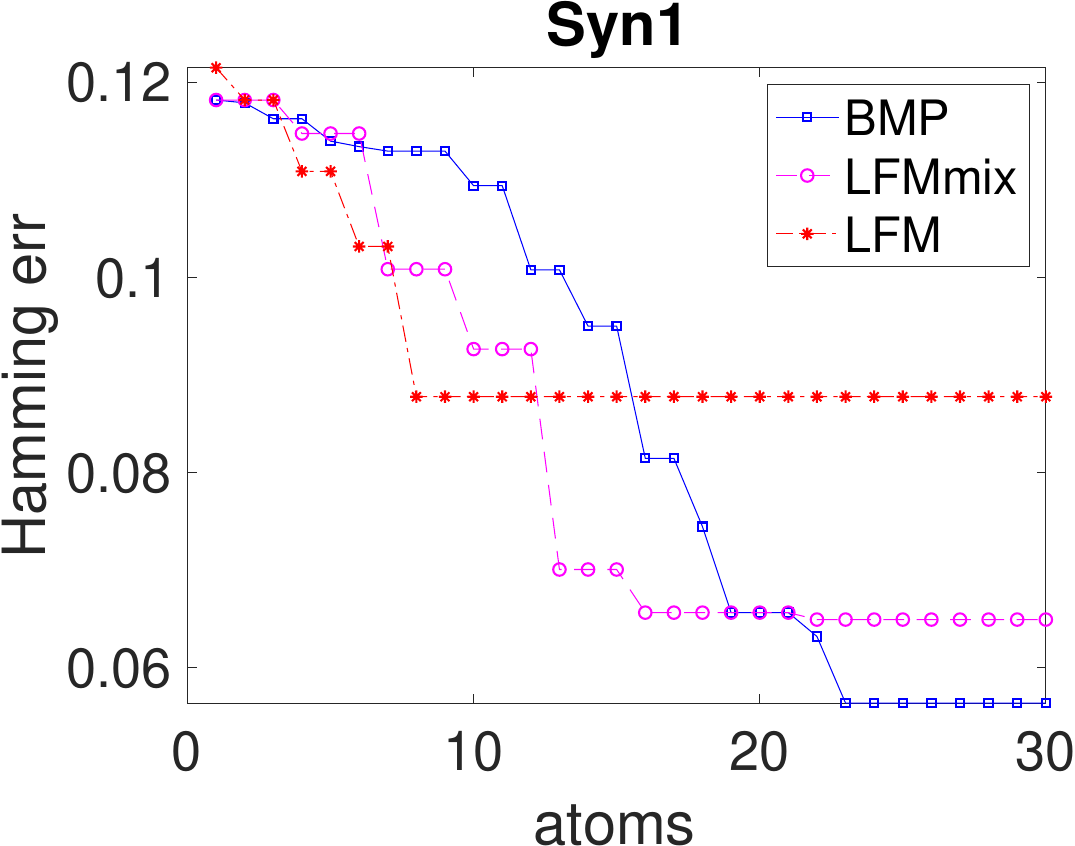}  }
\subfigure{ \includegraphics[width=0.30\textwidth]{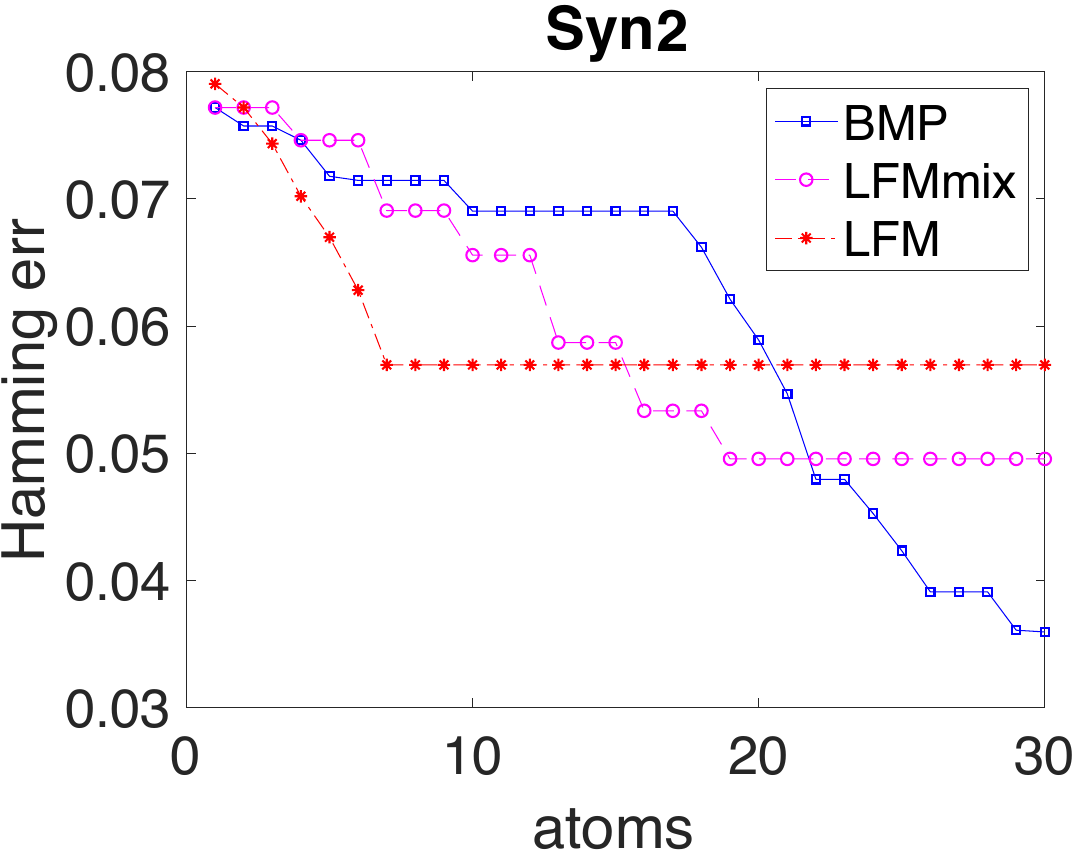}  }\\
\vspace{-4mm}
\subfigure{  \includegraphics[width=0.30\textwidth]{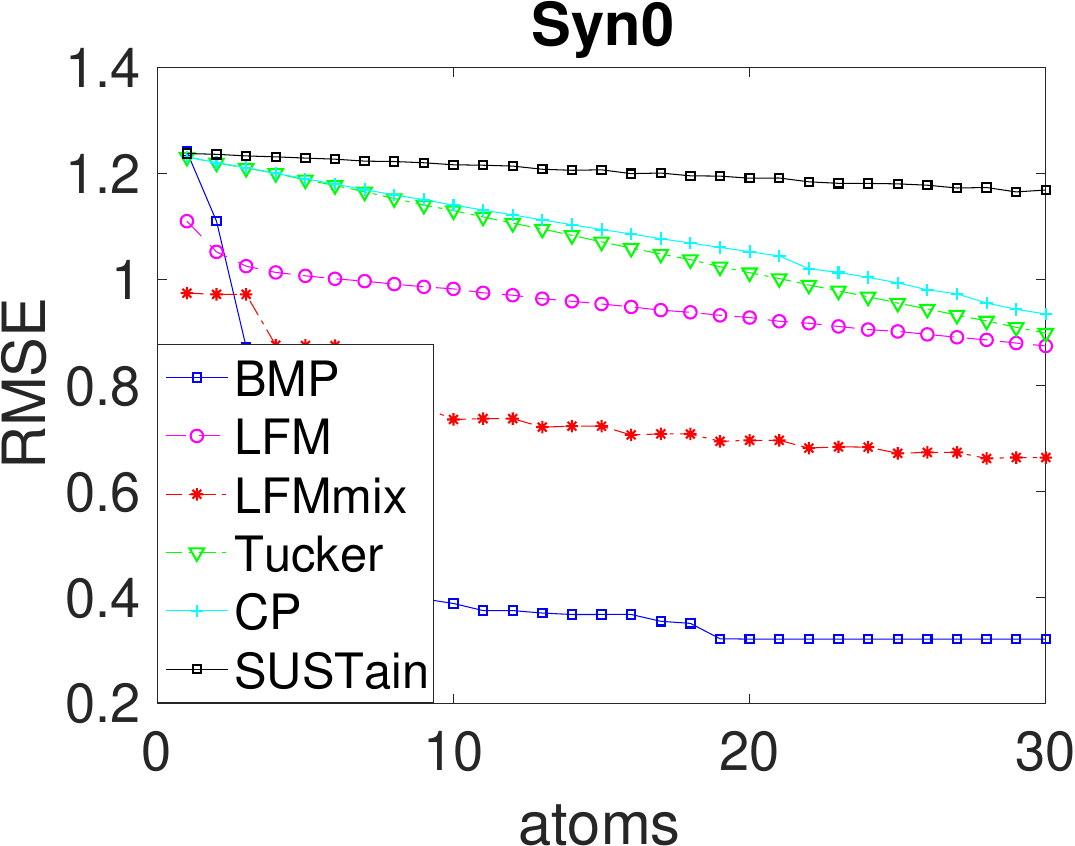}  }
\subfigure{  \includegraphics[width=0.30\textwidth]{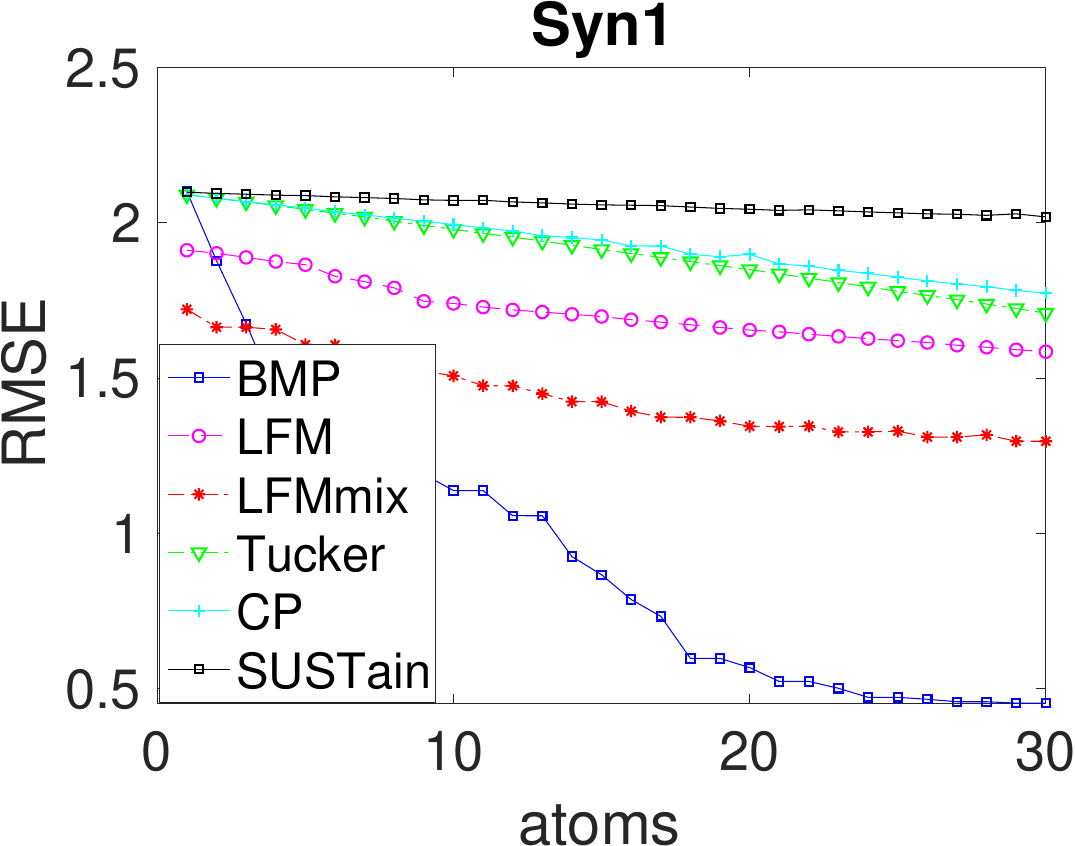}  }
\subfigure{  \includegraphics[width=0.30\textwidth]{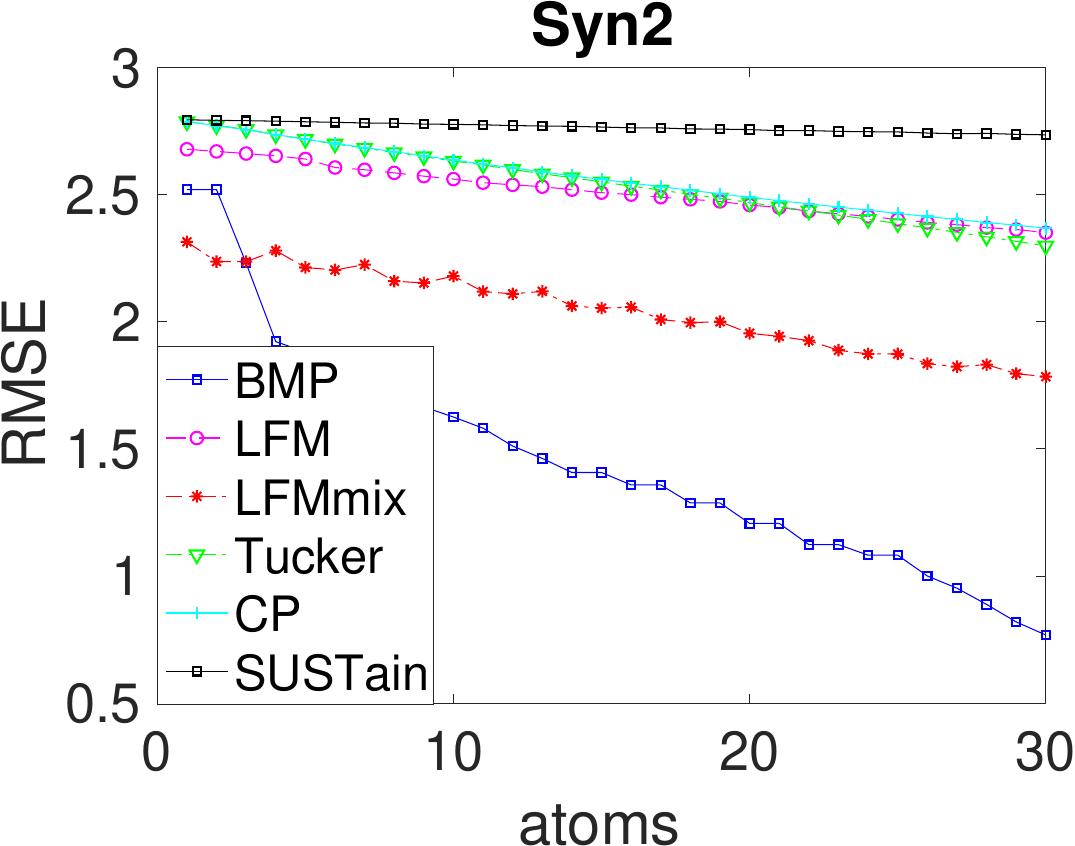} }
\vspace{-3mm}
\caption{Performance comparison w.r.t number of atoms on synthetic datasets. Top: denoising RMSE between the ground truth and the estimated tensor.  Middle: Hamming-error of the recovered Boolean factors. Bottom: completion RMSE between the ground truth and the estimated tensor.
}\label{fig:denoise_rmse}
\vspace{-5mm}
\end{figure*}

\section{Experiments}
\label{sec:exp}
The evaluate the performance of our algorithm, we experiment with two benchmark tensor decomposition tests: denoising and completion. We compare our method with the following baselines:
\begin{itemize}[noitemsep]
\item \texttt{LFM} \cite{yen2017latent}:  matrix factorization with Boolean constraints, solved for every mode separately. Results are reported from the best mode. 

\item  \texttt{LFMmix}: a mixture of latent feature matrix factors from \texttt{LFM}.

\item  \texttt{Tucker}: Tucker tensor decomposition solved with Higher Order SVD \cite{de2000multilinear}. 

\item  \texttt{CP}:  canonical polyadic tensor decomposition solved with ALS.

\item \texttt{SUSTain}~\cite{perros2018sustain}: hierarchical ALS tensor decomposition  with  Boolean projection. 
\end{itemize}

\vspace{-3mm}
\subsection{Synthetic Experiments}
\vspace{-3mm}
We randomly generate tensors  according to   model \eqref{eqn:model} of size $(150\times 150 \times 150)$. For every atom $ r \in [\bar{K}] $ and  mode  $n$, we generate a binary vector  $ \boldsymbol{z}^r_n \in \bool^{d_n} $ from a binomial distribution with probability $0.5$, and  vectors $\boldsymbol{u}^r_n,  \boldsymbol{w}_{n}^r$ from Gaussian. We vary the  number of atoms: $ \bar{K} \in \{3, 9, 15 \}$, and produce three synthetic datasets \textbf{Syn0}, \textbf{Syn1}, and \textbf{Syn2}.   
%

Tensor denoising aims to estimate the tensor $\T{W}$ from  observations $ \mathcal{X}$ contaminated by additive Gaussian noise. We synthesize the noise $\T{E} \sim \mathcal{N}(0,\V{1})$  and  evaluate the root-mean-square error (RMSE) between the ground truth tensor $ \T{W} $ and the estimate $\hat{\T{W}} $. Figure ~\ref{fig:denoise_rmse} top row shows the RMSE comparison of different methods  w.r.t number of atoms. \texttt{BMP} significantly outperforms the baselines. We also observe sublinear convergence rate as predicted by Theorem \ref{thm:convergence1}.   

To validate parameter recovery, we compare the Hamming distance between the estimated Boolean factors $\{ \V{z}^r_n \}$  and the ground truth, as shown in  Figure~\ref{fig:denoise_rmse} middle row. We can see that matrix-based methods \texttt{LFM} and \texttt{LFMmix} get stuck easily, and the recovery error stops decreasing after 10 atoms. In contrast, \texttt{BMP} can successfully recover most of the Boolean factors. Under noise, it is generally not possible to recover all the Boolean factors.


For tensor completion, a tensor $\T{X}$ has missing values and the task is to complete the missing entries solely based on observations. We consider noiseless completion and randomly remove $10$ percent of the entries from the ground truth tensor $\T{X}$.  Figure \ref{fig:denoise_rmse} bottom row shows the RMSE between the ground truth $\T{X}$ and the completed tensor $\hat{\T{W}}$ from decomposition. We can see that \texttt{BMP} achieves the lowest completion error with only a few numbers of atoms. On the other hand, \texttt{SUSTain} performs poorly as the  rounding step slows down the convergence.



\vspace{-3mm}
\subsection{Decoding Consciousness: Study on Brain Computer Interface}
\vspace{-3mm}
We study the neural mechanism underlying consciousness using large-scale neural activity  data recorded via ECoG at high temporal (>1 KHz) and spatial (3 mm) resolution \cite{yanagawa2013large} with $128$ electrodes.  The  data were recorded from the lateral cortex in macaques during rest, anesthetic and recovery conditions.  We form the data into a tensor of  (space $\times$ time $\times$ trial). Boolean factors indicate whether certain latent features appear at certain spatial locations, temporal positions or trials. We split the spatiotemporal ECoG recordings into 100 bins to simulate trials.  Each bin contains measurements over 3000 milliseconds.

\begin{figure*}[t]
\centering
\subfigure{ \includegraphics[width=0.225\textwidth]{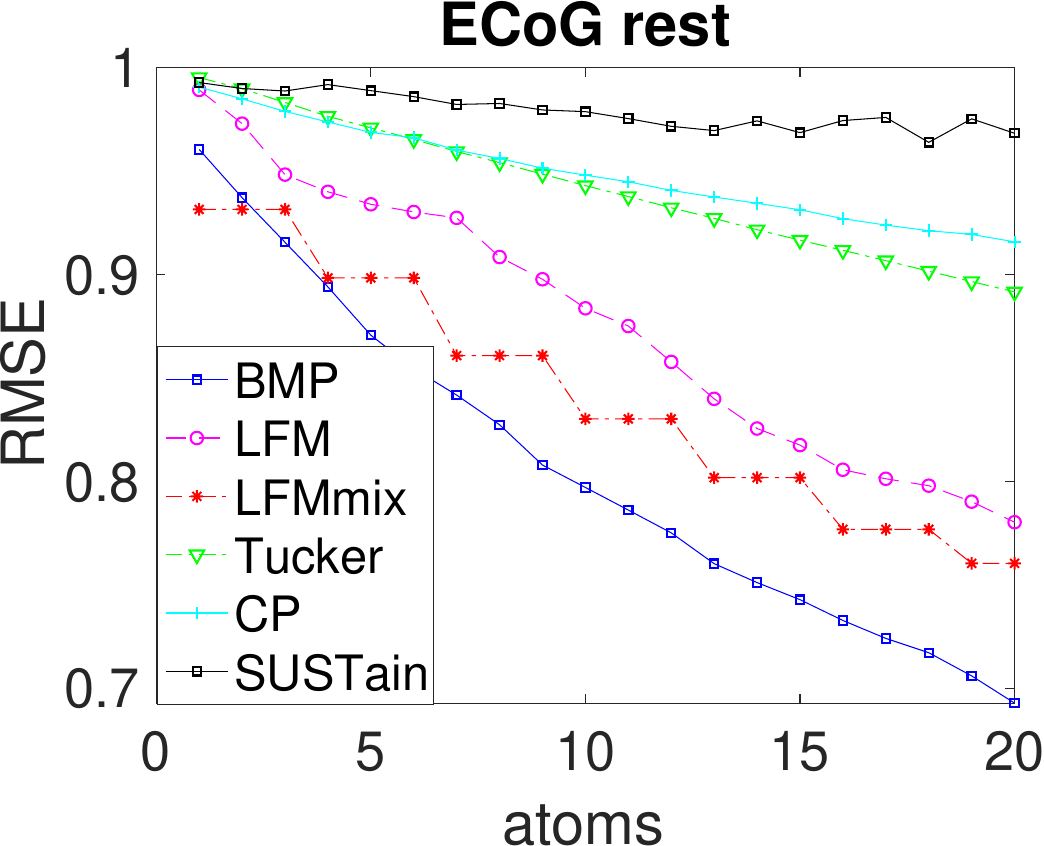}  }
\subfigure{ \includegraphics[width=0.225\textwidth]{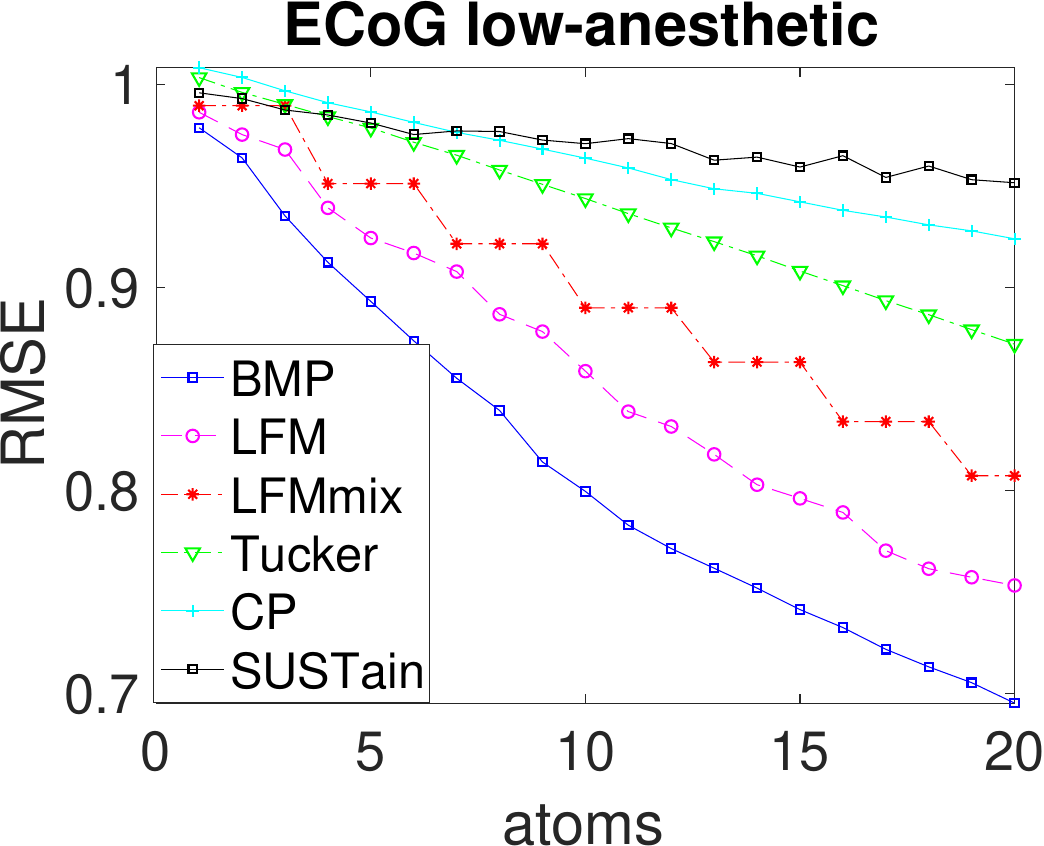}  }
\subfigure{ \includegraphics[width=0.225\textwidth]{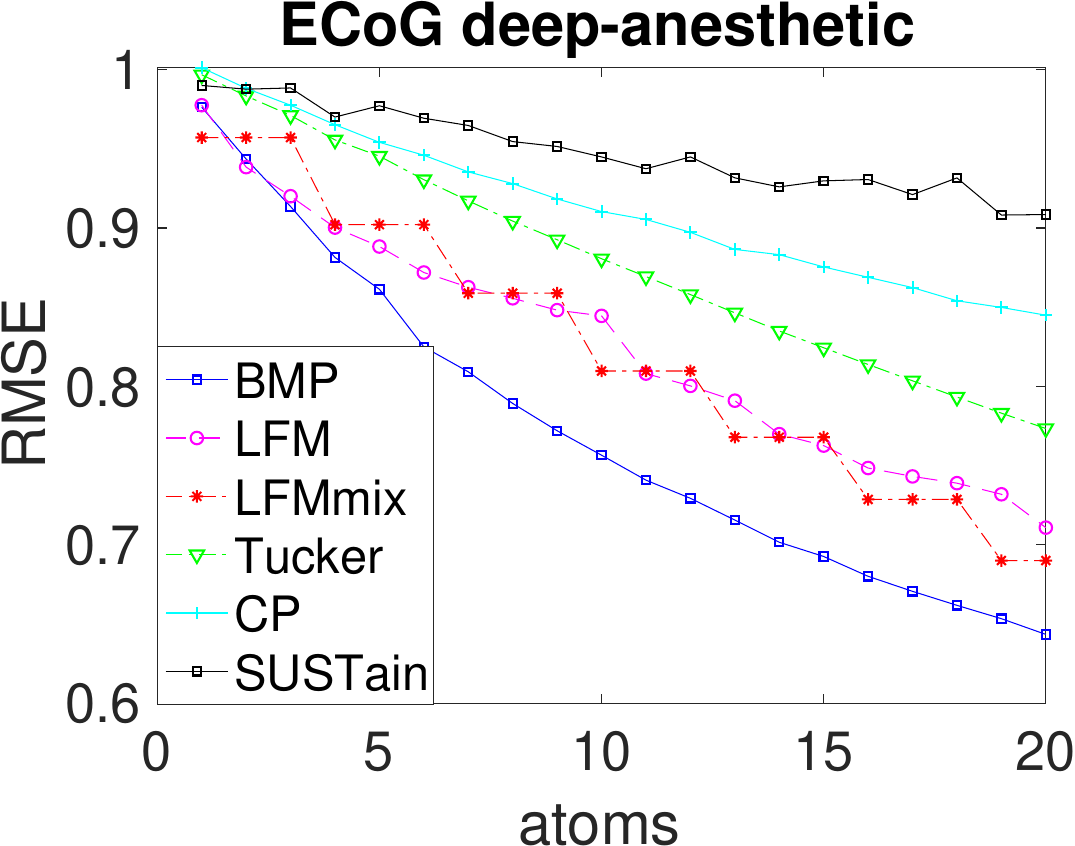}  }
\subfigure{ \includegraphics[width=0.225\textwidth]{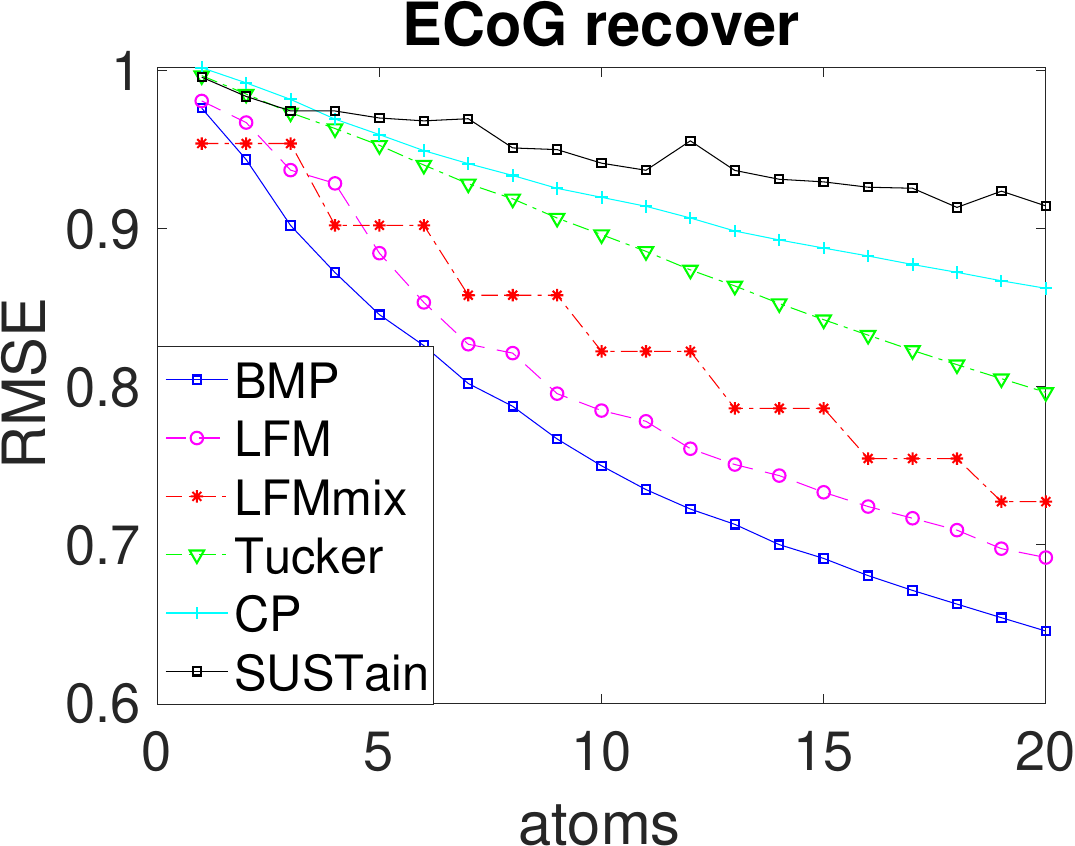}  }\\
\vspace{-4mm}
\subfigure{  \includegraphics[width=0.225\textwidth]{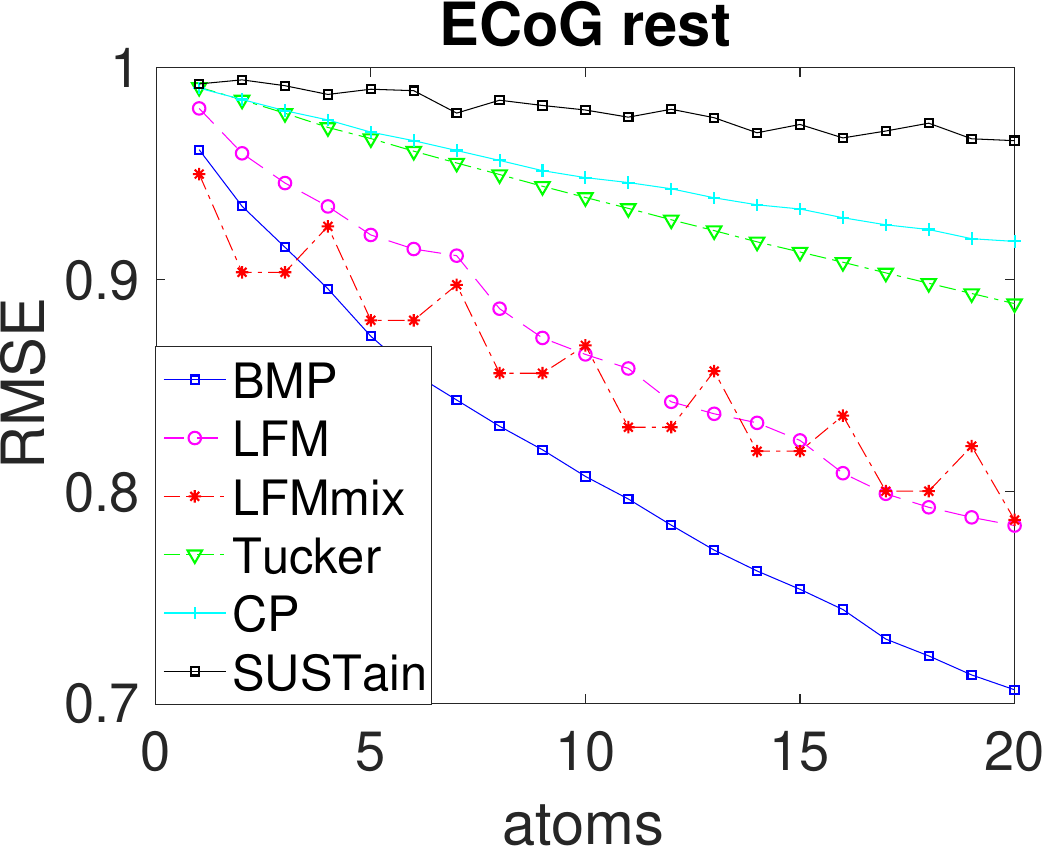}  }
\subfigure{  \includegraphics[width=0.225\textwidth]{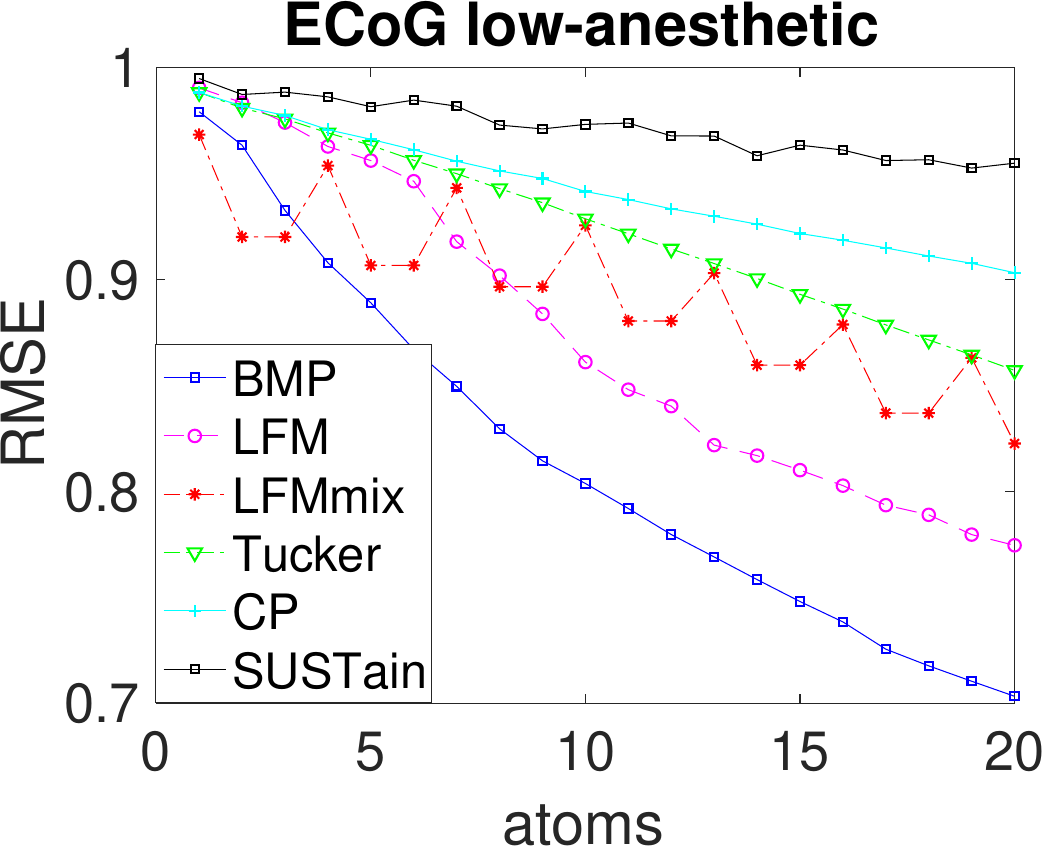}  }
\subfigure{  \includegraphics[width=0.225\textwidth]{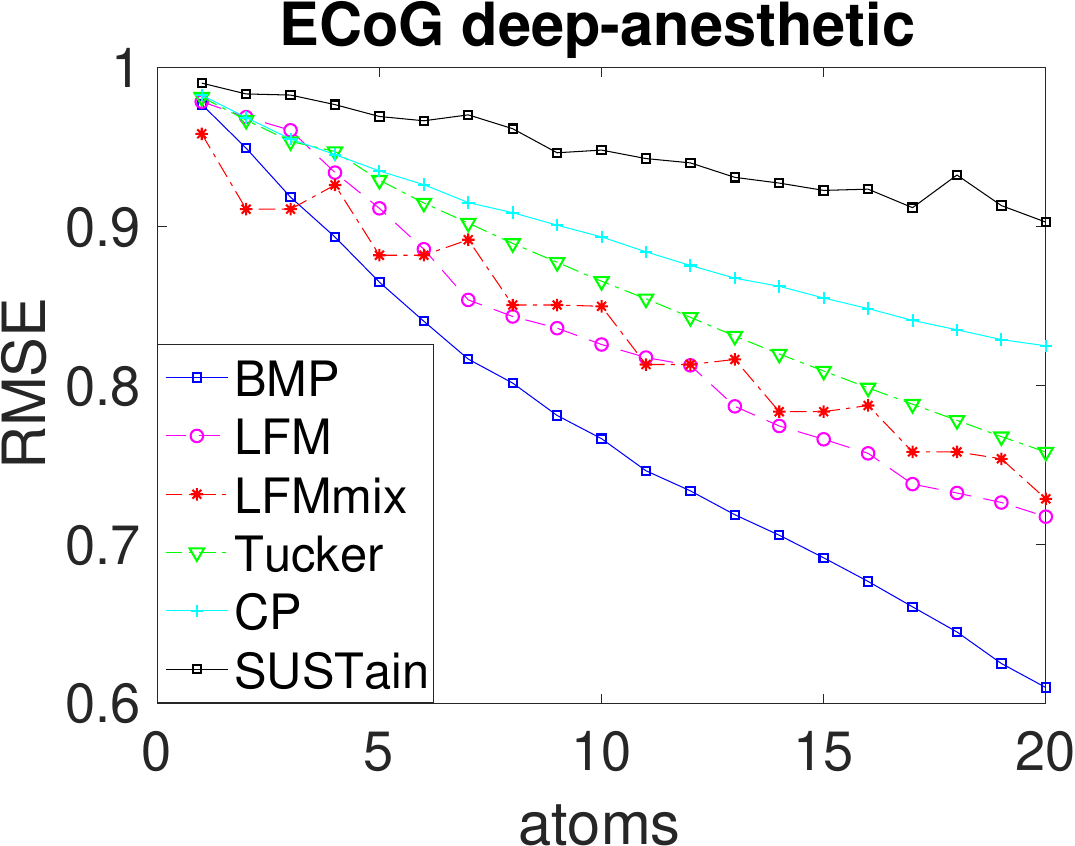}  }
\subfigure{  \includegraphics[width=0.225\textwidth]{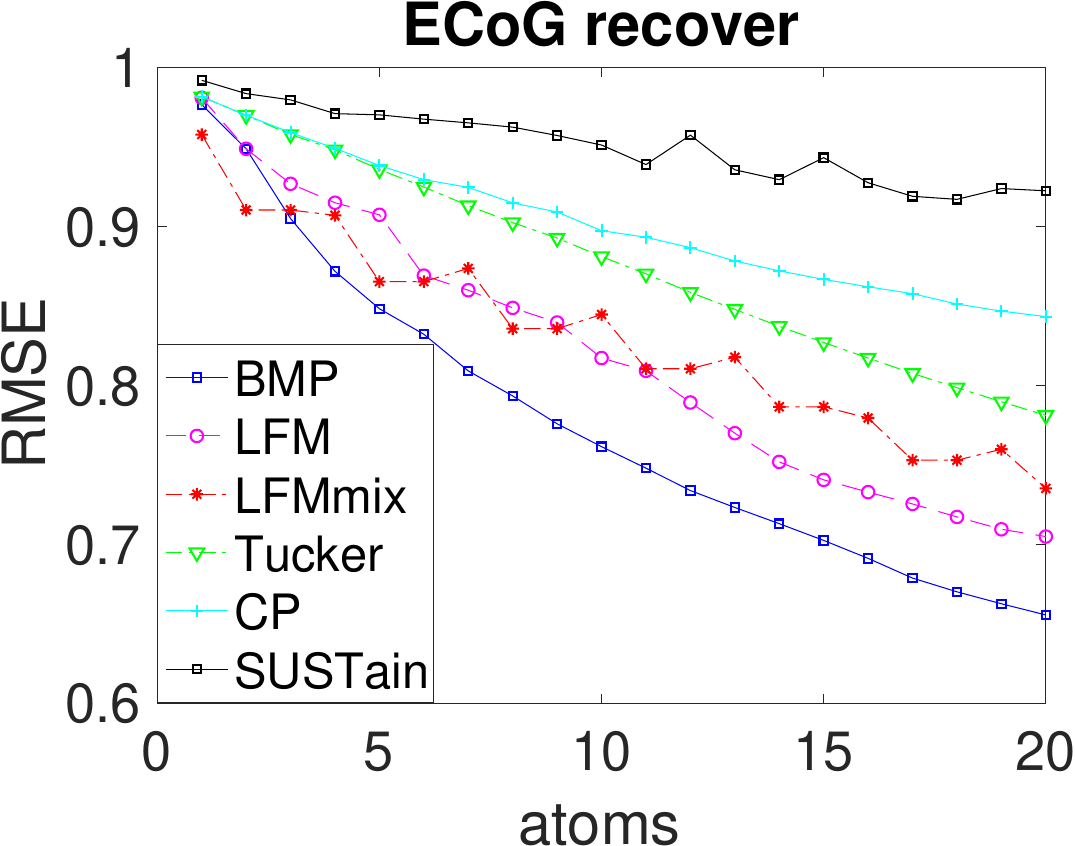}  }
\vspace{-2mm}
\caption{Performance comparison w.r.t number of atoms on real-world ECoG datasets for different methods. Top: denoising RMSE. Bottom: completion RMSE.
}\label{fig:recovery_rmse}
\vspace{-5mm}
\end{figure*}

We conduct  tensor denoising and completion experiments on the ECoG recordings following the same setting as the synthetic experiments. Figure \ref{fig:recovery_rmse} shows the RMSE comparison with varying number of atoms  for denoising (top row) and completion (botom row). \texttt{BMP} demonstrates clear advantages in these tasks, especially with only a  few number atoms.

\begin{figure}[!htbp]
\begin{minipage}[h]{0.32\textwidth}
{\centering\includegraphics[scale=0.18]{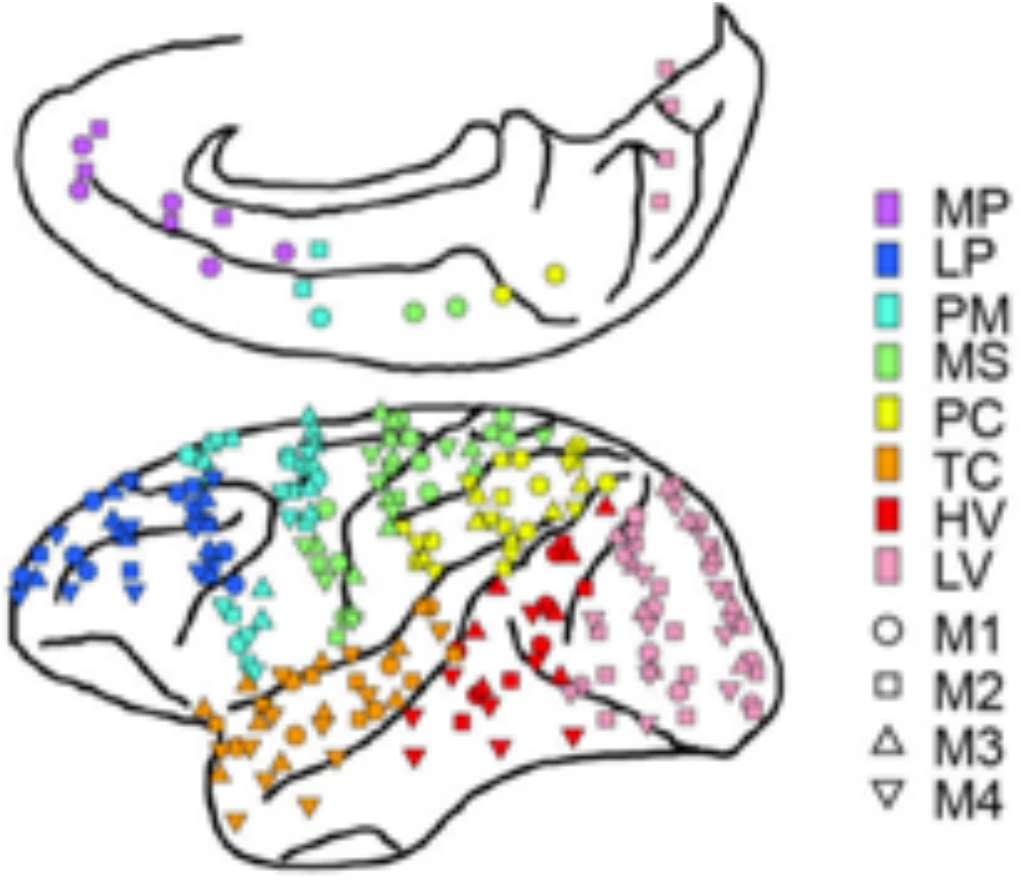}}
\caption{Bipolar ECoG electrode arrays. MP: Medial prefrontal cortex, LP: Lateral prefrontal cortex, PM: Premotor cortex, MS: Primary motor and somatosensory cortices, PC: Parietal cortex, TC: Temporal cortex, HV: Higher visual cortex and LV: Lower visual cortex.  }\label{fig:brain}
\end{minipage}
\hspace{2mm}
\begin{minipage}[h]{0.65\textwidth}
{\centering \includegraphics[scale=0.26]{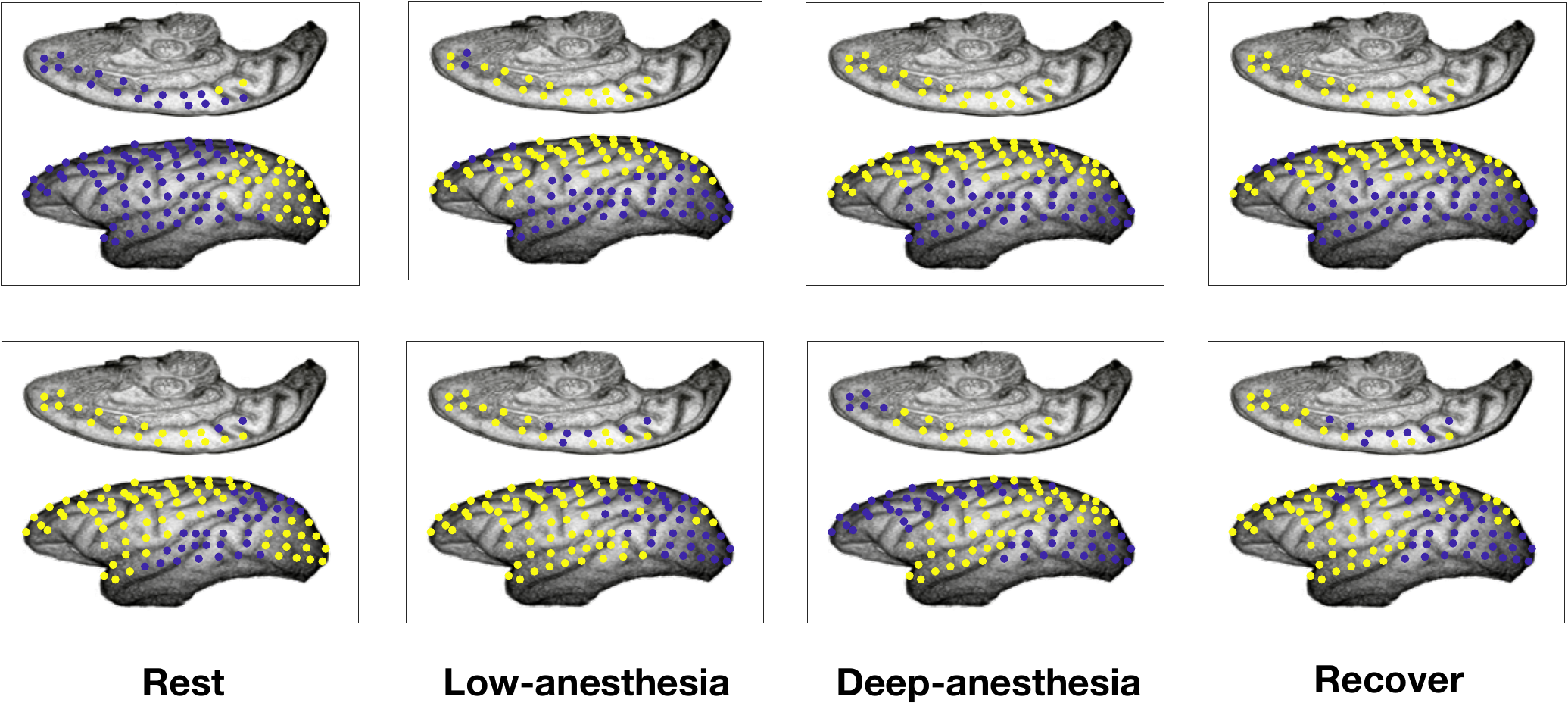}}
\vspace{-3mm}
\caption{Boolean factors of two atoms: atom 2 (top row) and atom 4 (bottom row) learn from BCI datasets corresponding to different monkey brain consciousness conditions. Factors are color coded (Yellow for 1 and blue for 0), indicating the presence or absence of  certain latent features.
}\label{fig:BCI}
\end{minipage}
\end{figure}

Figure \ref{fig:brain}  shows the bipolar re-referenced ECoG electrode arrays.  We visualize the learned Boolean factors (yellow for 1, blue for 0) on the spatial mode in Figure \ref{fig:BCI}. Interestingly, the learned factors have direct correspondence with brain anatomy. For example, In atom 2, Lower visual cortex (LV) is active in rest and deactivated under anesthetic, demonstrating visual consciousness. In atom 4,  Temporal cortex (TC) and Higher visual cortex (HV) is  deactivated while the monkey is resting. Low-anesthesia and recover share similar patterns. Deep anesthesia deactivates Lateral Prefrontal cortex which is  critically involved in broad aspects of executive behavioral control. The results demonstrate the power of our method in discovering underlying neural interactions.

\vspace{-4mm}

\section{Conclusion}
\label{sec:con}
We proposed an efficient optimization algorithm \texttt{BMP} for solving tensor decomposition with Boolean factors, leveraging atomic norm regularization and Boolean quadratic program. We proved that \texttt{BMP} can achieve sublinear convergence and polynomial run-time and sample complexity. We experimented exhaustively on synthetic and real-world  datasets and observed superior performance for both tensor  denoising and recovery tasks. When applied to ECoG recording, our method reveals the interesting neural mechanism underlying brain consciousness conditions.

\bibliographystyle{unsrt}
\bibliography{references}

\section{Appendix}
\label{sec:app}
\renewcommand{\thesubsection}{\Alph{subsection}.}

\subsection{Proof of Theorem 1}
Given an atomic set of rank-1 tensors $\mathds{A} = \bigcup \mathds{A}_n $, where $\mathds{A}_n  = \{\T{M}| \T{M} = \V{z}^r_n\otimes \V{u}^r_n\otimes \V{w}^r_n\}$, the span of an atomic set  is defined as the linear combination of the atoms in the set:
$$
span(\mathds{A})=\{\cW \mid \cW=\sum_{\cM_k \in \mathds{A}} \V{c}_k \cM_k,\; \bc\in\mathbb{R}^{|\mathds{A}|} \}
$$
The orthogonal projection of the set is:
\begin{align*}
\Delta \cW_{\bot} := \proj_{span(\mathds{A})^{\bot}}(\Delta \cW), \quad 
\cM_{\bot} := \proj_{span(\mathds{A})^{\bot}}(\cM),
\end{align*}
to denote projection onto the orthogonal space of the span of the active set of atoms $\mathds{A}$ at the beginning of the iteration. The atomic norm of a tensor $\T{W}$ with arbitrary Boolean factors is:
\[\| \mathcal{W} \|_{\mathds{A}}  = \big\{ \inf \ \| \boldsymbol{c} \|_1 : 
\mathcal{W} = \sum_{\mathcal{M}_k \in \mathds{A}} \V{c}_k \mathcal{M}_k,  \T{M} = \V{z}^r\otimes\V{u}^r\otimes \V{w}^r   \big\}\]
Let $\T{M}^\star$ be the atom selected by the Algorithm, the following lemma shows that minimization of a linear function with an atomic-norm regularization is achieved by minimization w.r.t. the greedy atomic direction.
\begin{lemma}\label{lemma:atom} 
Given a tensor $C$ and $\cW$, the following equality holds: 
\begin{align*}
&\min_{\cW\in \mathbb{V} }\; \langle C, \cW \rangle + \frac{\beta}{2}\| \cW \|_{\dA}^2 =\min_{\V{c}}\;  \langle C, \V{c}\cM^*_{\bot} \rangle + \frac{\beta}{2}\|\V{c}\cM^*_{\bot}\|_{\dA}^2 \\
\end{align*}
where $\mathbb{V}$ is some linear subspace, $\cM^*_{\bot}=\proj_{\mathbb{V}}(\cM)$, and $\V{c} \in \real^{| \mathds{A}|}$ is the weight vector.
\end{lemma}

\begin{proof}
The minimization problem with an atomic norm regularization
\begin{equation}\label{tmp2}
\begin{aligned}
\min_{\cW\in \mathbb{V} }\; \langle C, \cW \rangle + \frac{\beta}{2}\| \cW \|_{\dA}^2
\end{aligned}
\end{equation}
is equivalent to the constrained minimization problem
\begin{equation}\label{tmp3}
\begin{aligned}
&\min_{\cW\in \mathbb{V} }\;  &&\langle C, \cW \rangle 
&s.t.                         &&\|\cW\|_{\dA}\leq \delta
\end{aligned}
\end{equation}
for some constant $\delta$. Since the problem of form \eqref{tmp3} involves a linear objective function with a convex constraint, it always has an optimal solution that lies on the boundary of the atomic-norm constraint set. Therefore,  there is always a solution to \eqref{tmp2} of the form $\cW=c\cM^*_{\bot}$.
\end{proof}

Now we state the proof for Theorem 1.

\begin{proof}
Let $F$ be a $\beta$-smooth loss function w.r.t the solution $\T{W}$, its gradient is  $\beta$-Lipschitz, we have
\begin{equation}\label{tmp1}
\begin{aligned}
F(\cW')-F(\cW)  \leq \langle \nabla F(\cW), \cW' - \cW \rangle + \frac{\beta}{2}\|\cW' - \cW \|^2_F.
\end{aligned}
\end{equation}

At $(k+1)$-th iteration, we greedily add a new atom $\cM^* \gets \argmin_n \langle \nabla F(\cW), \cM_n \rangle$ into the active set $\mathds{A}^{k}$.
A key ingredient of our analysis is the $\mu = 3/5$
constant-approximation guarantee in the greedy step
given by the SDP-based MAXCUT solver, where we
have the following guarantee:
\[\langle \nabla F(\cW), \cM^* \rangle  \leq \min_n \langle  \nabla F(\cW), \cM_n  \rangle  \leq \mu \min_n\min_{\cM_n \in \mathds{A}_n} \langle  \nabla F(\cW), \cM_n  \rangle = \min_{\cM \in \mathds{A}} \langle  \nabla F(\cW), \cM  \rangle\]
After the fully-corrective weight adjustment in \eqref{adjustment},  we have
$$
F(\cW^{k+1}) = \min_{\cW\in span(\mathds{A}^{k+1})} F(\cW) \leq \min_{ c\in\mathbb{R} }\; F(\cW^{k}+ c \cM_{\bot}^*)
$$
for some constant $c$, given that $\cW^{k}+ c \cM_{\bot}^*\in span(\mathds{A}^{k+1})$. 
\begin{equation}\label{tmp1}
\begin{aligned}
F(\cW^{k+1})-F(\cW^{k}) 
 \leq & \min_{ c\in\mathbb{R} }\; F(\cW^{k}+ c \cM_{\bot}^*) -F(\cW^{k}) \\ 
 \leq & \min_{ c\in\mathbb{R} }\;  \langle \nabla F(\cW^k), c \cM_{\bot}^*\rangle   +  \frac{\beta}{2}\|c\cM^*_{\bot}\|_{\dA}^2 \\
\end{aligned}
\end{equation}

By lemma \ref{lemma:atom}, we have
\begin{align*}
\min_{c}\;  \langle \nabla F(\cW^k), c\cM^*_{\bot} \rangle + \frac{\beta}{2}\|c\cM^*_{\bot}\|_{\dA}^2 
&=\min_n\min_{\Delta \cW\in span(\mathds{A}^{k}_n)^{\bot} }\; \langle \nabla F(\cW^k), \Delta \cW \rangle + \frac{\beta}{2}\|\Delta \cW\|_{{\dA}^2_n}\\
\end{align*}
 Note that since $\langle \nabla F(\cW^{k}), \Delta \cW \rangle=0$ for any $\Delta W\in span(\mathds{A}^{k}_n)$, we also have
\begin{align*}
&\min_n\min_{\Delta \cW\in span(\mathds{A}^{k}_n)^{\bot} }\; \langle \nabla F(\cW^k), \Delta \cW \rangle + \frac{\beta}{2}\|\Delta \cW\|_{{\dA}^2_n} \\
&= \min_{\Delta \cW  \in span(\mathds{A})}\; \langle \nabla F(\cW^k), \Delta \cW \rangle + \frac{\beta}{2}\|\Delta \cW_{\bot}\|_{\dA}^2 .
\end{align*}
By constraining $\Delta W$ to be of the form $\Delta W = \alpha (\cW^*-\cW^k)$ , we have
\begin{align*}
&\min_n\min_{\Delta \cW \in span(\mathds{A}_n)^{\bot}  }\; \langle \nabla F(\cW^k), \Delta \cW \rangle + \frac{\beta}{2}\|\Delta \cW_{\bot}\|_{{\dA}^2_n}  \\
&\leq \min_{\Delta\cW=\alpha (\cW^*-\cW^k),\alpha\in[0,1] }\; \mu\langle \nabla F(\cW^k), \Delta \cW \rangle + \frac{\beta}{2}\|\Delta\cW_{\bot}\|_{\dA}^2 \\ 
&\leq \min_{\alpha\in[0,1] }\; \alpha\mu(F(\cW^*)-F(\cW^k)) + \frac{\beta\alpha^2}{2}\| \cW^*_{\bot}\|_{\dA}^2
\end{align*}
where the second inequality is from convexity. 
Let $F^*:=F(\cW^*)$, minimize over $\alpha$, we have
\begin{equation*}
\begin{aligned}
&(F(\cW^{k+1})-F^*)-(F(\cW^k)-F^*)\\
&\leq \min\{\frac{\mu^2}{2\beta\|\cW^*\|_{\dA}^2}(F(\cW^k)-F^*)^2, \frac{\mu}{2}(F(\cW^k)-F^*)\}
\end{aligned}
\end{equation*}
The recurrence then leads to the convergence result:
\[F(\cW^{k})-F^*  \leq \frac{2\beta \|\cW^*\|_{\dA}^2}{\mu^2}\frac{1}{k}\]
\end{proof}

\subsection{Proof of Theorem 2}

\begin{proof}
Denote the support set $\bar{A} \subset [\bar{K}]$, meaning $\V{c}_j=0$ for $j\notin \bar{A}$. Define  $\bc_{\bar{A}}:=\{\bc_{k}\}_{k \in\bar{A}}$ and $
f(\bc_{\bar{A}}):=F(\sum_{k \in\bar{A}} c_{k} \cM_k ).
$
Suppose $f(\bc_{\bar{A}})$ is strongly convex with respect to $\bar{A}$ with parameter $\gamma$.
Since $\bc^*$ is the minimizer of $f(\bc)$, it satisfies
$\langle \nabla f(\bc^*), \bc^* \rangle =0$ and thus
\begin{align*}
f(0)-f(\bc^*)
&=f(0)-f(\bc^*)-\langle \nabla f(\bc^*),0-\bc^*\rangle\\
&\geq \frac{\gamma}{2}\|\bc^*\|^2.
\end{align*}
Since, $f(0)-f(\bc^*)\leq \frac{1}{2}\|\cX\|_F^2$, we have 
\begin{equation}
\|\cW^*\|_{\dA}^2=\|\bc^*\|_1^2\leq \bar{K}\|\bc^*\|_2^2 \leq \frac{\bar{K}\|\cX\|_F^2}{\gamma}.
\end{equation}
Substituting \eqref{tmp3} into \eqref{convergence1} yields the result \eqref{convergence2}.
\end{proof}


\subsection{Proof of Theorem 3}
The following Kruskal's condition \cite{kruskal1977three} guarantees the uniqueness of generic tensor decomposition:
\begin{condition}[Kruskal]\label{app:kruskal}
An order-3 rank-R
tensor of dimension $d_1\times d_2\times d_3$ is unique if
\[R \leq \frac{1}{2}\left(\text{krank}(\M{Z}) + \text{krank}(\M{U}) + \text{krank}(\M{W}) -2 \right) \]
\end{condition}
where $\text{krank}(\cdot)$ is the largest value of a matrix  such that every subset of columns of the matrix is linearly independent. It is also easy to see that $\text{krank}(\M{Z}) \leq \text{rank}(\M{Z})$ for any $\M{Z}$.

To guarantee uniqueness of the Boolean factor $\M{Z}$, we need the following condition on its column vectors vectors $\{\V{z}_r\}$  to hold. 
\begin{condition} 
 \label{app:boolean}
The tensor decomposition problem has unique Boolean factors  if for any  non-trivial combinations of the column vectors $\{\V{z}_r\}$ would lead to a non-Boolean vector:
 \[ \forall \V{c}\neq \V{0}, \quad  \V{c}^\top \V{Z} \in \bool \iff \V{c} \in \{ \V{e}_i\}\]
\end{condition}
which means the linear subspace of $\V{Z}^n$ does not contain any other Boolean vectors that are not already in $\M{Z}^n$. This identifiability condition is in nature  similar to \cite{slawski2013matrix}.

A tensor decomposition $\T{W}= \sum_{r=1}^R  \V{z}_r \otimes \V{u}_r \otimes \V{w}_r$ is said to be unique if for any other decomposition, $\T{W}= \sum_{r=1}^R  \V{z}_r' \otimes \V{u}_r' \otimes \V{w}_r'$, there exists a permutation $\sigma$ such that 

\[ \V{z}_r \otimes \V{u}_r \otimes \V{w}_r = \V{z}_{\sigma(r)}' \otimes \V{u}_{\sigma(r)}' \otimes \V{w}_r'\]

When tensor $\T{W}$ has a unique decomposition, the set of vectors $\V{z}$ , $\V{u}$ and $\V{w}$ are identifiable, up to scalars.  Let $\M{Z}$ be the matrix containing $\V{z}_r$ as its columns. Similarly for matrices $\M{U}$ and $\M{W}$.

\begin{proof} Suppose all the continues-valued factors are normalized, with $\|\V{u}\|_2, \|\V{w}\|_2 \leq 1$. The optimization problem of  parameter recovery without noise  is equivalent to 
\begin{eqnarray}
\min_{\V{c}} \|\V{c}\|_1 \quad \text{s.t.} \quad \T{X} =\sum_{k=1}^{\bar{K}} \V{c}_k \T{M}_k, \T{M}_k = \V{z}_k\otimes \V{u}_k \otimes \V{w}_k
\label{eqn:lasso}
\end{eqnarray}
where $\bar{K}$ is the size of the atom set $\mathds{A}$. The problem in \eqref{eqn:lasso} has a unique solution $\V{c}^\star$, which  selects  the true atoms as $\{\T{M}_k^\star = \V{z}_k^\star\otimes \V{u}_k^\star \otimes \V{w}_k^\star\}$. Under the  identifiability conditions,  the tensor decomposition is unique up to scalars. For any  solution returned by the algorithm $\{\T{M} =\V{z}_k\otimes \V{u}_k\otimes \V{w}_k\}$,  there exists a permutation $\sigma$, such that $\V{u}^\star_k = \V{u}_{\sigma(k)}$ and $\V{w}^\star_k = \V{w}_{\sigma(k)}$. Given that the Boolean factors are unique by Condition \eqref{app:boolean}, $\T{M}_k$ is rank-1 and we can recovery $\V{u}_k$ and $\V{w}$ exactly. Therefore,  the linear subspace spanned by $\{\V{u}_k\}$  is the same as that of $\{\V{u}^\star_k \}$, similar for  $\{\V{w}_k\}$.

\end{proof}

\subsection{Proof of Theorem 4}

\begin{definition}[Restricted Strongly Convex (RSC)]
\label{thm:rsc}
For a given tensor $\T{W}$, we say the loss function $F(\T{W})$ is restricted strongly convex with parameter $\kappa$, if 
\[ \|\triangledown F(\T{W})\|_{op} \geq \kappa \|\T{W}\|_F^2\]
where $\|\cdot\|_2$ is the spectral norm and  $\|\cdot\|_F$ is the Frobenius norm. 
\end{definition}

It is easy to see that given the atomic set of rank-1 tensors, the atomic norm of the tensor $\|\T{W}\|_{\mathds{A}}$ is equivalent to the tensor nuclear norm. Consider the case where the tensor decomposition has orthogonal factors. That is  $\M{Z}^\top \M{Z} = \M{I}$  $\M{U}^\top \M{U} = \M{I}$ and $\M{W}^\top \M{W} =\M{I}$ obtained through orthogonalization, we have:
\[\|\T{W}\|_{\mathds{A}} = \{\inf \|\V{c}\|_1 \ | \ \T{W} = \V{c}_k \V{z}_k\otimes \V{u}_k \otimes \V{w}_k \} =  \{ \inf_{\V{\sigma}} \|\V{\sigma}\|_1   \ | \  \T{W} = {\sigma}_r \V{z}_k\otimes \V{u}_k \otimes \V{w}_k, \langle \V{z}_i, \V{z}_j \rangle =0 \}\]

Given the statistical estimation problem of sample size $S$:
\[\hat{\T{W}} =  \min_{\T{W}} \mathbb{E} [  \frac{1}{2S}F_i(\T{W}) + \lambda \|\T{W}\|_{\mathds{A}}]\]
where per sample loss $F_i(\T{W})$ is defined as 
$F_i(\T{W}) =  \|\mathfrak{X}(\T{X}_i - \T{W})\|_F$. $\mathfrak{X}$ is a linear operator representing vectorization or random sampling. Assume the true low-rank tensor parameter is $\T{W}^\star$, the statistical estimation difference $\hat{\Delta} :=  \hat{\T{W}}  -\T{W}^\star $ can be decomposed into two parts:  components that are in the span of $\mathds{A}$, denoted as $\Delta$ and the components that are in the complement of $\mathds{A}$, denoted as $\Delta_{\bot}$. As the atomic norm is decomposable w.r.t the subspace of $\mathds{A}$, we have  $\hat{\Delta} =  \Delta  + \Delta_{\bot}$ and $\|\hat{\Delta}\|_{\mathds{A}} = \|\Delta\|_{\mathds{A}}+ \|\Delta _{\bot}\|_{\mathds{A}}$. The following lemma relates  the atomic norm of these two components.

\begin{lemma}
Let the estimation error be $\|\mathfrak{X}(\hat{\Delta})\|_F = \|\mathfrak{X}(\hat{\T{W}} - \T{W}^\star)\|_F $ satisfying $\hat{\Delta} = \Delta + \Delta_{\bot}$, where $\Delta = \text{proj}_{\mathds{A}}(\hat{\Delta} )$ and $\Delta_{\bot}$ is the projection onto the complement of $\mathds{A}$. If the regularization $\lambda$ satisfy $\lambda \geq 2 \|\triangledown F(\T{W})\|_{\mathds{A}}$,  we have
$ \text{rank}(\Delta) \leq 2R $  and $\|\Delta_{\bot} \|_{\mathds{A}} \leq 3 \|\Delta\|_{\mathds{A}}$. 
\label{lemma:complement}
\end{lemma}

\begin{proof} Since the true model is low-rank $\T{W}^\star \in \mathds{A}$ based on our algorithm and the atomic norm $\|\cdot\|_{\mathds{A}}$ is decomposable w.r.t. the rank-1 tensors. Proof follows similarly from  Lemma 1 in \cite{negahban2012unified}.

By H\"older's inequality and triangle inequality, we have
\begin{eqnarray}
\frac{1}{2S}\|\mathfrak{X}(\hat{\Delta})\|_F^2 \leq \frac{1}{S}\langle \mathfrak{X}(\T{E}), \hat{\Delta}\rangle +\lambda_S \|\hat{\Delta} \|_{\mathds{A}}\leq \frac{1}{S} \|\mathfrak{X}(\T{E})\|_{op} \|\hat{\Delta}\|_{\mathds{A}} +\lambda_S \|\hat{\Delta} \|_{\mathds{A}} \leq 2\lambda_S \|\hat{\Delta} \|_{\mathds{A}}
\label{eqn:holder}
\end{eqnarray}
With the choice of $\lambda_S \geq \|\mathfrak{X}(\T{E})\|_{op}/S $.  By Lemma \ref{lemma:complement}, we have 
\begin{equation}
    \|\hat{\Delta}\|_{\mathds{A}}= \|\Delta\|_{\mathds{A}} + \|\Delta_{\bot} \|_{\mathds{A}}\leq 4 \|\Delta_{\bot} \|_{\mathds{A}} \leq 4\|\Delta\|_F \sqrt{2R}
\label{eqn:rsc}
\end{equation}
By Definition \ref{thm:rsc}, we have the lower bound $\frac{1}{2S}\| \mathfrak{X}(\Delta)\|_F^2 \geq  \kappa(\mathfrak{X})\| \Delta \|_F^2  $ with $\kappa(\mathfrak{X})$ as the RSC constant of the linear operator $\mathfrak{X}$. Combining Eqn. \eqref{eqn:holder} and \eqref{eqn:rsc}, we have $\|\hat{\Delta}\|_F \leq  8\lambda_S \sqrt{2R}/\kappa $.

For i.i.d Gaussian noise $\T{E}  \sim \mathcal{N}(0,\sigma^2)$, apply concentration bound. There exists a universal constant $c_1$ such that:
\begin{eqnarray}
\|\mathfrak{X}(\T{E})\|_{op}^2  \leq  \mathcal{O}_p \big(c_1 \sigma^2R(d_1 + d_2+ d_3)\big).
\label{eqn:noise}
\end{eqnarray}
Combine Eqn. \eqref{eqn:holder}, \eqref{eqn:rsc} and  \eqref{eqn:noise}   together, the following statically bound  for the estimation error holds for some constant $c_1$ with high probability: 
\[\|\mathfrak{X}(\hat{\T{W} }  - \T{W}^\star)\|_F^2   \leq \mathcal{O}_p \big(c_1 \frac{\sigma^2R(d_1 + d_2+ d_3)}{S}\big) \]
which is proportional to the noise variance and the degree of freedom in the tensor model.
 \end{proof}

\end{document}